\documentclass[11pt]{article}
\usepackage[margin=1in]{geometry} 

\usepackage{amsmath}

\usepackage{subcaption}
\usepackage{multirow}
\usepackage{graphicx}
\usepackage{tabularx}
\usepackage{adjustbox}
\usepackage{multirow}
\usepackage{caption}
\usepackage{subfiles}
\usepackage[numbers]{natbib}

\newcolumntype{P}[1]{>{\centering\arraybackslash}p{#1}}

\makeatletter
\renewcommand\@fnsymbol[1]{\the#1}
\makeatother

\begin{document}

\title{When Automated Assessment Meets Automated Content Generation: Examining Text Quality in the Era of GPTs}

\author{
Marialena Bevilacqua\footnote{University of Notre Dame, Indiana, USA \{mbevilac@nd.edu, koketch@nd.edu, rqin@nd.edu, wstamey@nd.edu, xzhang38@nd.edu, aabbasi@nd.edu\}} 
\and
Kezia Oketch\footnotemark[1] 
\and
Ruiyang Qin\footnotemark[1] 
\and
Will Stamey\footnotemark[1] 
\and
Xinyuan Zhang\footnotemark[1] 
\and
Yi Gan\footnote{Georgia Tech, Atlanta, Georgia, USA \{yi.gan@scheller.gatech.edu\}} 
\and
Kai Yang\footnote{Shenzhen University, Shenzhen, China \{kayang6-c@my.cityu.edu.hk\}} 
\and
Ahmed Abbasi\footnotemark[1] 
}

\date{\today}
\maketitle








\begin{abstract}
The use of machine learning (ML) models to assess and score textual data has become increasingly pervasive in an array of contexts including natural language processing, information retrieval, search and recommendation, and credibility assessment of online content. A significant disruption at the intersection of ML and text are text-generating large-language models such as generative pre-trained transformers (GPTs). We empirically assess the differences in how ML-based scoring models trained on human content assess the quality of content generated by humans versus GPTs. To do so, we propose an analysis framework that encompasses essay scoring ML-models, human and ML-generated essays, and a statistical model that parsimoniously considers the impact of type of respondent, prompt genre, and the ML model used for assessment model. A rich testbed is utilized that encompasses 18,460 human-generated and GPT-based essays. Results of our benchmark analysis reveal that transformer pretrained language models (PLMs) more accurately score human essay quality as compared to CNN/RNN and feature-based ML methods. Interestingly, we find that the transformer PLMs tend to score GPT-generated text 10-15\% higher on average, relative to human-authored documents. Conversely, traditional deep learning and feature-based ML models score human text considerably higher. Further analysis reveals that although the transformer PLMs are exclusively fine-tuned on human text, they more prominently attend to certain tokens appearing only in GPT-generated text, possibly due to familiarity/overlap in pre-training. Our framework and results have implications for text classification settings where automated scoring of text is likely to be disrupted by generative AI.

\end{abstract}

\section*{Keywords}
Automated Essay Scoring, auto-generated text, user-generated content, text quality, generative AI, large language models



\section{Introduction}

The use of machine learning (ML) models to quantify or “score” textual data has become increasingly pervasive over the past 25-30 years. In natural language processing (NLP), there are a myriad of text sequence classification problems related to categorization of text based on topics, sentiment, affect, and psychometric dimensions \cite{liang2023few,ahmad2020deep,cheng2023consistent}. In information retrieval (IR), text-based documents are scored for relevance assessment in various application and system contexts, including search relevance for search engines \cite{sakai2022relevance,shao2023understanding} and recommendation engines used in recommender systems \cite{wu2023personalized,qin2023automatic}. The uptick in digital content and activity has been accompanied by the rise of illicit digital content. Accordingly, ML-based scoring has also become important for credibility assessment in an array of online settings including detection of web spam, phishing websites, deceptive reviews, and fake news \cite{abbasi2012detecting,liu2020fned}.

Transformers \cite{devlin2018bert,vaswani2017attention}, and foundational large-language models (LLMs) such as generative pre-trained transformer (GPT) \cite{radford2019language,brown2020language}, represent a major disruption at the intersection of ML and text, affording unprecedented opportunities for ML-based assessment and generation of text \cite{TOISStylizedDataText, thorp2023chatgpt}. Nonetheless, much of the evidence of LLM performance on traditionally human processes and tasks is anecdotal first-person (experiential) accounts, often disseminated through social media and blog postings. There is a nascent yet emerging body of literature highlighting GPT’s performance on various tasks. Some of the most robust early evidence of capabilities comes from assessments such as licensing and graduate coursework exams \cite{kung2023performance, terwiesch2023would,singhal2023Nature} and other curriculum activities \cite{ drori2022neural}.

There remains a need for rigorous empirical and experimental research on the implications of generative AI in a bevy of contexts. The potential for man-machine hybridization recently received scrutiny across an array of activities \cite{fugener2022cognitive,berente2021managing,seidel2018autonomous}. Accordingly, \textit{the research objective of this study is to empirically assess the impact of hybrid environments on ML-based text assessment/scoring, where both the assessment and generation of textual content involve a mixture and/or interplay between humans and ML.} Alternatively stated, rather than exploring the effectiveness of ML generated content directly against a human gold standard, as done in emerging literature \cite{drori2022neural,terwiesch2023would}, we explore differences in how ML-based scoring models trained on human content assess the quality of content generated by humans versus GPTs. 

The specific context we use to explore this gap is automated essay scoring (AES). In AES, traditionally, ML-models are used to score human-generated essays \cite{Ramesh2022-kg,ijcai2019p879}. There are several AES contexts where such ML-based assessments happen. Most notably in educational settings, but also potentially in talent analytics, personnel selection, and job-to-candidate recommender systems \cite{sajjadiani2019using,zhao2021embedding}. In educational testing, ML-models are used to assess essays in K-12 education and other tests such as the writing portions of the GMAT, GRE, and TOEFL \cite{rudner2000overview,shermis2002exit}. AES provides a suitable experimental setting due to the abundance of publicly available human-generated essays, expert ratings, and accompanying prompts well-suited for GPT-based essay generation. Further, although there is limited benchmarking of ML methods for AES, the state-of-the-art resembles methods used in related ML-based text scoring research such as sentiment, user experience ratings, psychometrics, and personality \cite{liang2023few,ahmad2020deep,yang2023getting}. Our three research questions are as follows:

\begin{itemize}
	\item
	\textbf{RQ1:} How effective are state-of-the-art feature-based and deep learning models for AES?
	\item 
	\textbf{RQ2:} How do AES models trained on human-generated content rate text generated by GPT models? What is the moderating effect of different document genres on such assessments?
	\item
	\textbf{RQ3:} What linguistic categories and cues are most different for human versus GPT-generated text?
\end{itemize}

To address these questions, we propose a rich analysis framework that encompasses state-of-the-art ML-models for essay scoring, human and ML-generated essays with a prompt engineering protocol for the latter, and a statistical model that holistically considers essay genre types, human versus machine authorship, and the scoring model to parsimoniously infer main effects and interactions. We couple our framework with a rich testbed encompassing 15,437 human-generated and 3,023 GPT-based essays (1,537 GPT-3.5 and 1,486 GPT-4) associated with 68 prompts related to 6 document genres. Results of our benchmark analysis (RQ1) reveal that transformer language model-based methods such as BERT and RoBERTa more accurately score human essay quality as compared to CNN/RNN and feature-based ML methods, attaining scoring mean squared errors that are 10-40 percent lower. Interestingly, in regards to RQ2, we find that the transformer language model-based methods also tend to score GPT-generated text 10-15\% higher on average, relative to documents authored by humans. This is interesting because the GPT data is out-of-sample for these models trained on human-only text. Conversely, the traditional deep learning and feature-based ML models score human-generated text considerably higher than GPT text. Moreover, these effects are most pronounced for certain types of document genres such as narrative, argument, and response. Further analysis related to RQ3 reveals that the transformer-based scoring models attend to certain tokens appearing in GPT-generated text more prominently.

Our main contributions are two-fold. First, we develop an analysis framework for contexts involving automated content assessment and generation mechanisms, including ML scoring models, GPT-generated texts based on prompt design adaptation, and parsimonious statistical models for evaluating such intersections. We intend to make all benchmarking code, analysis models, and prompt design processes publicly available. Second, we use AES to offer various empirical insights, including how best-in-class text scoring methods, based on transformer language models, may score certain genres of GPT-generated content higher even when exclusively trained on human content, possibly due to familiarity. That is, overlap in language modeling data sources and transformer-based attention head mechanisms. Although set in the context of AES, our framework and results have implications for many text classification and IR settings where automated scoring of text/documents is likely to be disrupted by generative AI in the coming years, often in unintended and uninvited ways, including search relevance (e.g., of web pages, blogs, social media content), content recommendation, online credibility assessment, anti-aliasing, and fake news detection, just to name a few. The results of our work may be especially alarming for adversarial settings where generative models could disrupt the traditional profit-formula for return-on-fraud by alleviating barriers to entry for the creation of high-quality automated content.      

The remainder of this paper is organized as follows. In the ensuing section, we discuss relevant prior work related to AES. We then delve into the effectiveness of generative models such as GPT for human ability tasks, and review the state-of-the-art for ML-based scoring of human-generated content. In Section 3, we describe our analysis framework and testbed. Section 4 presents results for our benchmark evaluation of ML models for AES, as well as our statistical model-based empirical results for how ML scoring models rate human versus GPT text. Section 5 uses text content analysis and language model visualization tools to shed light on the underlying mechanisms driving our results. In Section 6, we offer concluding remarks.

\section{Related Work}
In this section we review work on automated essay scoring (AES), large-language models (LLMs) and text quality, and ML methods for scoring text.

\subsection{Automated Essay Scoring (AES)}
Essay assessment entails evaluating quality of textual content generated by human authors. It is viewed as a tool for assessing one’s ability to retain knowledge, synthesize ideas, interpret data, and express oneself through written language \cite{valenti2003overview, brown2017assessment}. Automation of essay assessment through AES was motivated by similar objectives as other ML-based text scoring use cases such as topic categorization and sentiment analysis: limited bandwidth and availability of expert raters, large volume of text needing assessment, timeliness, subjectivity of human judges, and so forth. The traditional process of scoring typically relies on the subjective judgement of fallible human assessors who are overworked, and often underpaid\cite{shermis2003automated, page1968use}. Moreover, the sheer volume of essays in these assessments has led to the implementation of AES to handle the workload as a co-rater, alleviating the sole reliance on human assessors\cite{dikli2006overview}. Although traditional AES approaches pre-date the rise of ML \cite{page1966imminence}, the underlying computational technology has evolved from rule-based scoring to ML models \cite{shermis2002exit}, including feature-based and deep learning ML models. 

Examples of popular AES systems are Project  Essay  Grader™ (PEG),  Intelligent Essay   Assessor™ (IEA), and E-rater® \cite{rudner2000overview}. In educational contexts, Massive Open Online Course (MOOC) organizations such as edX, MIT and Harvard's MOOC federation, utilize automated scoring \cite{balfour2013assessing}. Beyond the practical constraints necessitating AES, the reliability of AES is also often deemed on par with or better than human assessments due to training on a large set of documents that adds consistency in ratings relative to sole reliance on a large group of individual assessors that may yield high variance \cite{dikli2006overview}. Despite the widespread growth and adoption of AES, it is not without its detractors. Some are hesitant to incorporate AES systems into their organizational settings due to the differences in scores that machines may produce \cite{bridgeman2012comparison}. In regards to MOOCs, in contrast to edX’s extensive use of AES, Coursera, a Stanford-based MOOC, adheres to the traditional human assessment of essays \cite{balfour2013assessing}. Presently, benchmarking of ML methods for AES is limited in terms of the methods evaluated, the testbeds employed, and the evaluation metrics utilized \cite{ijcai2019p879,Ramesh2022-kg}. 

Our key research gap is as follows. \textit{Given our interest in exploring the interplay between ML-based assessment and generative AI versus human text, an important prerequisite research gap becomes performing an in-depth benchmark analysis of the performance of state-of-the-art ML methods for AES on multiple testbeds.} 

\subsection{Large Language Models and Text Quality}

The NLP space has made tremendous strides in the word embedding and language modeling space in the past 10 years, from static embeddings to contextualized embeddings and transformer-based large language models (LLM) \cite{mikolov2013distributed,brown2020language,anil2023palm}. Perhaps the most prominently featured and discussed amongst these LLMs are GPTs such as ChatGPT and other transformer-based models such as Llama and PaLM-2 \cite{openai2023gpt4,touvron2023llama,anil2023palm}. While a majority of research concerning the implications of such models is still to be conducted, initial success is already seen in such LLMs’ capabilities on knowledge tasks. In taking the United States Medical Licensing Exam (USMLE), which consists of three separate tests, ChatGPT performed "at or near the passing threshold" and  displayed both consistency among its responses and comprehension of the presented topic  \cite{kung2023performance}. More recent LLM benchmarks place their performance on such exams as being even higher \cite{singhal2023Nature}. Similar findings concerning ChatGPT's abilities are also seen in its performance on an MBA course's final exam \cite{terwiesch2023would}. ChatGPT3 was successful in passing the exam with a B/B- grade due to its Operations Management knowledge and explanations, in addition to its ability to correct itself after receiving human hints. While it was successful in passing, ChatGPT did not receive a higher grade due to its simple mathematical mistakes and inability to handle advanced process analysis questions. Conversely, the Codex LLM designed for code-based language modeling is capable of generating and solving university-level math problems \cite{drori2022neural}. ChatGPT has also seen some initial success in taking law school exams. Researchers at the University of Minnesota Law School discovered that ChatGPT achieved a passing grade on four different course exams, and may be considered a "mediocre" law student that could attain a JD degree from a reputable law school \cite{choi2023chatgpt}. Moreover, the PaLM-2 model can pass written proficiency exams for Chinese, Japanese, Italian, French, and Spanish \cite{anil2023palm}. ChatGPT was even recently used to assess the quality of essays, though it's assessment performance lagged behind that of feature-based machine learning methods \cite{mizumoto2023exploring}.

Outside of formal examination contexts, much of the evidence and discussion of generative AI’s ability to produce quality content in knowledge tasks, relative to human-generated content, remains anecdotal and underexplored. There is ample conjecture of the potential impact, opportunities, and challenges associated with LLMs in an array of contexts and occupations including within academia \cite{lund2023chatting}, \cite{king2023conversation}, \cite{liebrenz2023generating}. On the positive side, some believe LLMs may act as a proponent of interactive learning \cite{baidoo2023education}.

Text quality is a critical component of AES, which measures an essay's overall success in delivering its intended message. Text quality has been the focus of much research in various fields including NLP. Researchers have explored the factors that contribute to high-quality text, such as coherence, cohesion, lexical diversity, and grammatical accuracy \cite{crossley2020linguistic}. They have also investigated the cognitive processes involved in producing high-quality text, such as planning, revising, and self-evaluation \cite{tillema2011relating}. In the context of automated assessment and content generation, research has focused on the effectiveness of automated tools for evaluating and generating high-quality text. ML algorithms have been used to evaluate the quality of student writing and provide feedback to learners \cite{zhai2023effectiveness,bernius2022machine}. Similarly, NLP techniques have been used to generate text that is stylistically consistent with a particular writer's voice \cite{ficler2017controlling}. The dynamics of text quality assessment and generation become more crucial when introducing ML generated content into a human-ML generation-assessment process, particular with the advent of sophisticated models such as GPTs. Although such language models are equipped with skills to author logical and semantically accurate writings, they may need help creating compelling, subtle or enthusiastic narratives \cite{dathathri2019plug,radford2019language}. Human-authored essays, on the contrary, are inclined to have these humanistic tendencies because they are produced by people who are more knowledgeable and experientially attached with the subject matter, or at the very least, possess personal views and opinions. 

The main research gap we explore is the following. \textit{Given the nascent and emergent nature of studies that evaluate the capabilities of generative LLMs such as GPTs, and the confluence of ML-based scoring/assessments, there remains a need for studies that offer in-depth empirical evidence of the effectiveness of such LLMs in terms of quality of generated text across an array of prompt/genre types.}

\subsection{Machine Learning (ML) Methods for Automated Assessment of Text}
In this section, we review ML methods commonly used to score sequences of text, with specific emphasis on literature related to assessment of text quality and/or traits manifesting in (human) user-generated content. Relevant prior methods can be broadly grouped into three distinct categories: feature-based methods, deep learning convolutional and recurrent neural networks (CNN/RNN), and transformer-based pretrained language models (PLMs). Details are as follows.

\subsubsection{Feature-based ML}~\\
Until recent years, the mode for automated NLP research focusing on text categorization problems was manual feature engineering approaches. These features were typically combined with supervised ML classifiers such as k-nearest neighbors (KNN), support vector machines (SVMs), and gradient boosted trees \cite{pratama2015personality,tadesse2018personality}. For instance, predefined features such as number of words in a document, average word length, and number of spelling errors, that were input to a ML algorithm \cite{amorim2018automated}. The effectiveness of these models is heavily influenced by the choice and quality of features used for training \cite{yannakoudakis-etal-2011-new,alikaniotis2016automatic,uto2021learning, taghipour-ng-2016-neural}. Feature-based machine ML models, such as KNN, SVR, and XGBoost, used in conjunction with lexicons such as the linguistic inquiry and word count (LIWC), other domain-specific lexicons, and lexical measures related to word, sentence, and paragraph composition have shown promise in several text quality and trait scoring tasks \cite{yang2023getting,ijcai2019p879}. A study by \cite{bin2008automated} involved transforming essays into a vector space model and employing TF-IDF (term frequency inverse document frequency) and information gain for feature selection from words, phrases, and arguments. Training the KNN algorithm with the different feature selection methods led to a precision rate exceeding 76 percent on the CLEC corpus. Approaching automated scoring as a regression task, \cite{cozma2018automated} proposed an SVR-based model which combines histogram intersection string kernels and bag-of-super-word-embeddings features. They attained better performance on the AES task compared to other feature ML approaches. SVM and SVR have also been used in automated scoring studies to quantify scores for content and linguistic features input into the models \cite{peng2010automated,farnadi2013recognising,tadesse2018personality}. Similarly, XGBoost, an eXtreme Gradient Boosting algorithm, has garnered significant attention due to its exceptional performance in various data-driven tasks\cite{chen2016xgboost}. In contexts relevant to our study, XGBoost attained high accuracies for scoring text quality \cite{salim2019automated} and traits manifesting in the text \cite{tadesse2018personality}.

\subsubsection{Deep learning CNN/RNN Methods}~\\
Beyond the feature-based machine learning approach, deep learning techniques have emerged as a new paradigm in automated essay scoring. While traditional automated essay scoring systems rely on carefully designed features to evaluate and score essays\cite{uto2021learning,amorim2018automated} deep learning techniques, on the other hand, use end-to-end representation learning \cite{alikaniotis2016automatic,taghipour-ng-2016-neural,jin-etal-2018-tdnn,uto2020neural}. Recent studies have focused on neural network approaches because of their ability to garner enhanced text assessment performance \cite{shibata2022analytic}, giving better results compared to statistical models with handcrafted features\cite{dong2016automatic}. Both Convolutional Neural Networks (CNN) \cite{kim-2014-convolutional} and recurrent neural networks (RNN) such as long-short term memory (LSTM) and gated recurrent units (GRU) networks have been used to automatically score input text (including quality and related trait scoring tasks), typically in conjunction with word embeddings \cite{yu2017deep,yang2023getting}. For instance, several studies have employed single-layer LSTM over word embeddings for AES \cite{alikaniotis2016automatic,taghipour-ng-2016-neural}. Similarly, \cite{berggren2019regression} demonstrated that using pre-trained word embeddings along with GRUs enhanced the model's ability to understand the semantic nuances of essays, leading to improved AES accuracy. In regards to CNNs, 1D convolution over embedding vectors of text token sequences have been employed in a myriad of studies \cite{dong2016automatic,yu2017deep,majumder2017deep}.

\subsubsection{Transformer-based Pre-trained Language Models (PLMs)}~\\
Text scoring and sequence classification has seen significant advancement with the emergence of transformer-based pre-trained language models (PLMs) such as BERT, RoBERTa, and GPT. These models have revolutionized various NLP tasks and have shown promise in enhancing the accuracy and efficiency of automated assessment of text quality and traits. BERT (Bidirectional Encoder Representations from Transformers), a pretrained language model, has been widely adopted in various NLP applications due to its ability to capture contextual information from both left and right contexts in a sentence\cite{devlin2018bert,rodriguez2019language}. It has achieved state-of-the-art results in various NLP tasks\cite{kenton2019bert}. BERT has been applied to AES\cite{uto2020neural}, analysis of language competencies \cite{schmalz2021automatic}, automated short-answer assessment \cite{sung2019improving}, and inference of traits from textual data \cite{yang}, providing state-of-the-art performance in many cases. BERT-based assessment models excel in handling complex literary devices and where a nuanced understanding of the topic is beneficial \cite{rodriguez2019language}. RoBERTa (A Robustly Optimized BERT Pretraining Approach)\cite{liu2019roberta} builds upon the architecture of BERT by further optimizing the pre-training process via greater hyperparameter tuning, inclusion of additional training data, and more parameters \cite{liu2019roberta}. For instance, BERT uses 3.3 million tokens from the BookCorpus and Wikipedia. RoBERTa also uses BookCorpus and Wikipedia, but in addition to common crawl news (CC-News), open web text, and stories, resulting in over 30 million tokens used for pre-training \cite{liu2019roberta}. Both BERT and RoBERTa have performed well on text quality assessment and trait scoring tasks due to their ability to capture rich contextual information and linguistic nuances present in user-generated text \cite{ dong2021automated, gupta2023data, yang2023getting}.

The key research gap we tackle is as follows. \textit{With advancements in generative capabilities due to LLMs, in the context involving automated scoring of text using ML, it is unclear how various types of ML assessment methods designed for and trained on human content might assess GPT-based text; and how commonalities/differences between assessment models and those used to generate text might factor into the human versus GPT scoring dynamics.}

\section{Proposed Research Design and Analysis Framework}

\begin{figure}[ht]
        \centering
        \includegraphics[width=\columnwidth]{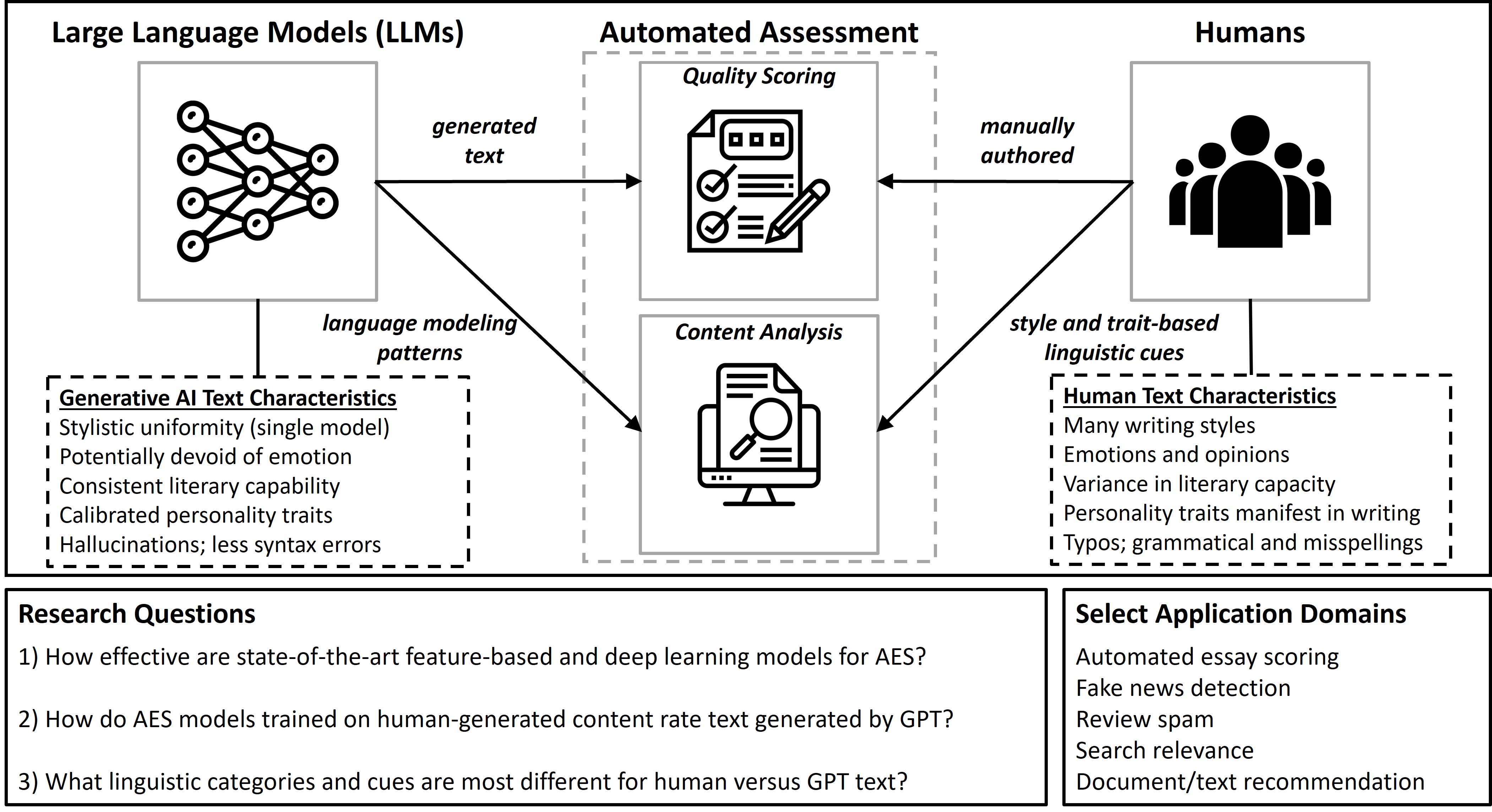}
        \caption{Overview of Research Design}
        \label{fig:researchDesign}
\end{figure}

Guided by the research gaps identified and our three research questions, Figure \ref{fig:researchDesign} presents an overview of our research design. As shown in the figure, we wish to use ML-based automated assessment of text (top center) as the grounds for exploring the interplay between human and LLM-generated text (top right and left in Figure \ref{fig:researchDesign}, respectively). Human text encompasses many stylistic and trait-based tendencies. These include digital traces of authors' syntactic, semantic, and lexical stylistic preferences and proclivities \cite{abbasi2008writeprints}, as well as their choices regarding conveyance of emotions and opinions \cite{liang2023few,cheng2023consistent} in authored texts. Moreover, human text also engenders variability in literary capabilities and likelihood of misspellings and grammatical mistakes \cite{ahmad2020deep}. Furthermore, as noted in our review of ML methods for scoring text, human texts also encompass personality traces \cite{yang2023getting,yang2020cnn,majumder2017deep}. Conversely, text generated by LLMs is likely to encompass greater stylistic uniformity, consistency in literary capabilities, and less emotion and syntactic mistakes \cite{radford2019language,dathathri2019plug}. Moreover, hallucinations remain a concern \cite{azamfirei2023large}, and persona-calibrated LLM text generation remains an open avenue of inquiry.

The contrast between human and LLM generated text is further moderated by the use of ML to assess such text. As noted in the introduction, we imagine generative AI content being injected into a myriad of processes and scenarios where ML models are currently scoring purportedly (human) user-generated content. In order to answer our three research questions, using AES as our focal testbed and evaluation setting, we propose automated assessment benchmarking, quality scoring, and content analysis (top center of Figure \ref{fig:researchDesign}. We begin by benchmarking quality of scores for ML models (RQ1), and then use statistical models to examine the interplay between human and LLM text using different ML models for scoring, across document genres and prompt types (RQ2). Finally, we employ information theoretic content analysis methods to shed light on the underlying mechanisms (RQ3) that may be responsible for observed effects from our statistical models.

\subsection{Analysis Framework}

\begin{figure}[ht]
        \centering
        \includegraphics[width=\textwidth]{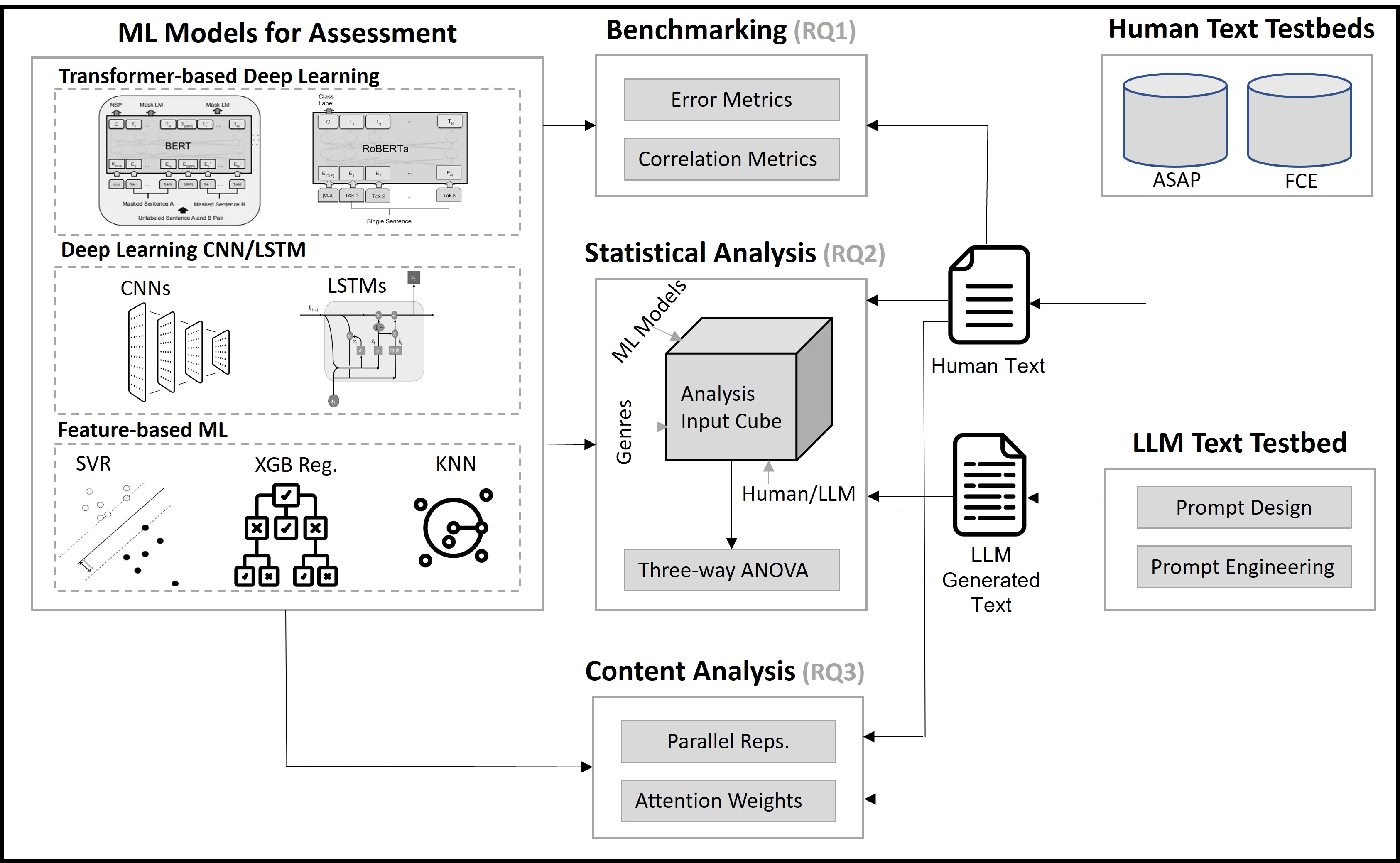}
        \caption{Overview of Analysis Framework}
        \label{fig:analysisFrame}
\end{figure}

Figure \ref{fig:analysisFrame} presents our proposed analysis framework. Consistent with prior surveys of state-of-the-art methods \cite{ijcai2019p879,Ramesh2022-kg,yang2023getting}, we incorporated three types of ML methods for scoring/assessing text: transformer-based deep learning, traditional CNN/LSTM deep learning methods, and feature-based ML techniques. Two types of testbeds were incorporated -- existing data sets comprising human-written essays and an LLM text testbed generated using ChatGPT. The ML assessment models were applied to the two types of text, resulting in three types of analysis corresponding to our three research questions: benchmarking, statistical analysis, and content analysis. Details about the proposed Analysis Framework are as follows.

\subsection{Testbed Overview}
\subsubsection{Human Text Testbeds}
The human text testbed included user generated text from two corpora: (1) the Automated Student Assessment Prize (ASAP); (2) the Cambridge Learner Corpus-First Certificate in English (CLC-FCE). ASAP\footnote {https://www.kaggle.com/c/asap-aes} was developed by the Hewlett Foundation with the goal of advancing the state-of-the-art for AES by consolidating and evaluating ML-based AES innovations. As depicted in the top-half of Table \ref{tab:dataset}, ASAP is comprised of eight essay sets. Each set varies based on the genre/prompt type of essay, writer grade level, dataset size, average length of essays in words, and quality score range. In total, we included the 12,977 essays from the training set segment of ASAP, because these are the ones comprising score labels. In regards to the prompts or genres of the essays, ASAP essay sets encompass three genres: argumentative, response, and narrative. The language levels of these writers range from US grades seven to ten. ASAP is considered a large dataset due to the number of essays per prompt and the resulting total number of essays \cite{ijcai2019p879}. The score range differs among the eight essay sets; some ranges are more narrow, such as essay sets three through six, whereas others are wider as seen in essay sets seven and eight. Since its inception, the ASAP dataset is a commonly used corpus for holistic scoring, in which the quality of an essay is represented by one score \cite{ijcai2019p879}. Due to the creation and widespread utilization of this dataset, a number of advancements have been made in ML-based AES techniques and models \cite{ taghipour-ng-2016-neural, phandi-etal-2015-flexible, jin-etal-2018-tdnn}. Accordingly, we incorporate it for our human essay testbed. 

CLC-FCE emerged from an amalgamation of the CLC project conducted by Cambridge University Press and Cambridge Assessment, as well as the First Certificate in English (FCE) exam. The CLC portion corresponds to essay sets 1-4 in Table \ref{tab:dataset}. The testbed also contains prompts and responses from the FCE, which is used to evaluate one's English at a higher level of learning. For these, authors must complete responses to one of two sets of prompts in which they are asked to write either an article, letter, report or short story. These are denoted by the last two rows in Table \ref{tab:dataset} corresponding to essay sets 5a and 5b. In total, CLC-FCE includes 2,460 texts spanning five different genres of prompts: argumentative, comment, letter, narrative, and suggestion. Responses to the prompts are between 200 and 400 words long, and scores ranging from 0-5 along a continuous scale are provided for each task. CLC-FCE has also been used extensively in prior AES benchmarking studies \cite{Ramesh2022-kg,ijcai2019p879}, thereby warranting inclusion in our study.

\begin{table*}[h]
    \small
    \centering
    \begin{tabular}{ |P{1.4cm}|P{1.4cm}|P{2.3cm}|P{1.1cm}|P{1.5cm}|P{1.7cm}|P{1.3cm}|} 
                \hline
                \textbf{Dataset} & \textbf{Essay Set} & \textbf{Type} & \textbf{Grade Level} & \textbf{Dataset Size (Essays)} & \textbf{Average Length (Words)} & \textbf{Scoring} \\
                \hline
                ASAP  & 1 & ARG & 8 & 1,785 & 350 & 2-12 \\
                \hline
                ASAP  & 2 & ARG & 10 & 1,800 & 350 & NA\\
                 \hline
                ASAP  & 3 & RESP & 10 & 1,728  & 150 & 0-3\\
                \hline
                ASAP  & 4 & RESP & 10 & 1,772  & 150 & 0-3\\
                \hline
                ASAP  & 5 & RESP & 8 & 1,805 & 150 & 0-4\\
                \hline
                ASAP  & 6 & RESP & 10 & 1,800 & 150 & 0-4\\
                \hline
                ASAP  & 7 & NARR & 10 & 1,730  & 250 & 0-30\\
                \hline
                ASAP  & 8 & NARR & 7 & 918 & 650 & 0-60\\
                \hline
                FCE & 1 & LETT & NA & 10 & 200-400 & 0-40\\
                \hline
                FCE & 2 & ARG, COMM, NARR, SUGG & NA & 10 & 200-400 & 0-5\\
                \hline
                FCE & 3 & ARG, COMM, LETT, NARR & NA & 10 & 200-400 & 0-5\\
                \hline
                FCE & 4 & ARG, COMM, LETT, NARR & NA & 10 & 200-400 & 0-5\\
                \hline 
                FCE & 5a & ARG, COMM, LETT, SUGG & NA & 10 & 200-400 &  0-5\\
                \hline
                FCE & 5b & ARG, COMM, LETT & NA & 10 & 200-400 & 0-5\\
                \hline
    \end{tabular}
    \caption{Overview of Human Text Testbeds}
    \label{tab:dataset}
\end{table*}

Prior information retrieval and information systems literature has underscored the importance of document genres \cite{yoshioka2001genre,auramaki1988speech,chang1994speech}. Genres manifesting in user-generated text provide insight into actions and intentions \cite{yoshioka2001genre,auramaki1988speech}. The ability to generate argumentative versus commentary or suggestion styles of text has important implications for sense-making and computational intelligence \cite{abbas2018text}. In AES, different prompt types elicit content corresponding to genres of text. In total, the human text testbed was comprised of 68 prompts related to the aforementioned six text genres. Table \ref{tab:prompts} provides examples of the six genres of prompts manifesting across the ASAP and CLC-FCE testbeds.   

\begin{table}[h]
    \small
    \centering
    \begin{tabular}{|P{1.3cm}|P{1.7cm}|P{1.3cm}|P{9.5cm}|}
                \hline
                \textbf{Dataset} & \textbf{Prompt ID} & \textbf{Genre} & \textbf{Prompt} \\
                \hline
                AES & 2 & ARG & "All of us can think of a book that we hope none of our children or any other children have taken off the shelf. But if I have the right to remove that book from the shelf -- that work I abhor -- then you also have exactly the same right and so does everyone else. And then we have no books left on the shelf for any of us." --Katherine Paterson, Author.
Write a persuasive essay to a newspaper reflecting your views on censorship in libraries. Do you believe that certain materials, such as books, music, movies, magazines, etc., should be removed from the shelves if they are found offensive? Support your position with convincing arguments from your own experience, observations, and/or reading. \\
                \hline
                FCE & 58 & COMM & You have had a class discussion on shopping.
Your teacher has now asked you to write a composition, giving your opinions on the following statement.

Shops should be open 24 hours a day, seven days a week.\\
                \hline
                FCE &  45 & LETT & You recently spent two days at an annual international arts festival.
Most of the events you went to were good, but you feel that the festival could be even better next year.\\
                \hline
               AES &  8 & NARR & We all understand the benefits of laughter. For example, someone once said, “Laughter is the shortest distance between two people.” Many other people believe that laughter is an important part of any relationship. Tell a true story in which laughter was one element or part.\\
                \hline
                AES & 5 & RESP & Describe the mood created by the author in the memoir. Support your answer with relevant and specific information from the memoir.\\
                \hline
                FCE & 22 & SUGG & A group of American students has just arrived in your town and the group leader has asked for information on an interesting building to visit.
Write a report for the group leader, describing one building and giving reasons for your recommendation.\\
                \hline
    \end{tabular}
    \caption{Example Prompt Types Associated with Six Genres in Human Text Testbeds}
    \label{tab:prompts}
\end{table}

\subsubsection{LLM Text Testbed}

Consistent with prior studies on the use of ChatGPT, a prompt design and engineering process was used to develop the LLM text testbed \cite{zheng2023can}. A prompt design and engineering team comprised of six members went through a series of meetings to discuss ideal strategies for prompting. Initially, the prompters took an inventory of the full set of prompts spanning the ASAP and CLC-FCE human essay testbeds. This resulted in a total of 68 unique prompt IDs associated with the six genres depicted in Table \ref{tab:prompts}. Given some of the CLC-FCE prompts, namely 5a and 5b in Table \ref{tab:testbed}, provide authors with up to 5 choices, the team decided it would be best to make each choice a separate prompt. After a few meetings, the team finalized the ChatGPT prompts with the goal of making them as similar to the human prompts as possible \footnote {Prompts and ChatGPT generated text  available at:  github.com/nd-hal/automated-ML-scoring-versus-generation}. This expansion resulted in approximately 150 prompts associated with the 68 prompt IDs. The team decided to use zero-shot learning to better align the LLM task with the human text generation process. Hence, each prompt was provided to ChatGPT as new conversations. Moreover, each prompt was provided to ChatGPT 10 times, resulting in 1537 total documents in the LLM text testbed. Some of the prompts required respondents to share opinions and experiences. ChatGPT often provides a disclaimer before generating responses to such prompts. The team elected to exclude any disclaimer text. These 1537 texts were generated using GPT-3.5. However, a follow-up collection constructed a second GPT testbed using the same prompts and procedures, but with GPT-4. Notably, ChatGPT powered by GPT-4 elected to not respond to two of the 68 prompts on account of "I'm sorry for any misunderstanding, but as of my training data cut-off in September 2021, there is no book titled "[Book title]" by [Publisher] specifically available or documented." Hence, the GPT-4 version of the ChatGPT testbed encompassed 66 prompts and 1486 total documents. 

\subsection{ML Models for Assessment}
As described in our review of methods used for automated assessment, prior work has mostly leveraged three types of ML methods \cite{ijcai2019p879,Ramesh2022-kg}: feature-based classifiers in conjunction with manually crafted, domain-specific features; word embeddings in coupled with CNN/RNN models; transformer-based PLMs fine-tuned on training data. Accordingly, representative methods from all three categories were incorporated in our set of ML models used for benchmarking and ML predictions input into the statistical analysis model. Details are as follows.  

The feature-based ML methods were ones used extensively in prior text classification work, including research that  examined textual traits of individuals \cite{yang2023getting}, as well as text quality \cite{Ramesh2022-kg}. In particular, three models were incorporated: XGBoost \cite{tadesse2018personality, salim2019automated}, SVM \cite{cozma2018automated, farnadi2013recognising}, and KNN \cite{bin2008automated, farnadi2013recognising}. Following prior work, XGBoost \cite{tadesse2018personality}, SVM \cite{farnadi2013recognising} and KNN \cite{farnadi2013recognising} all used Linguistic Inquiry and Word Count (LIWC) \cite{pennebaker2007linguistic}, a series of language-oriented lexicons, and lexical measures related to sentence/word lengths, number of sentences, etc., as input features. The occurrence of lexicon tags and individual items were included as binary presence vectors (i.e., “1” if present in that text, “0” if absent). Hyperparameters were tuned on the training data. For XGBoost, the XGB tree-based regressor model was used to run each iteration, with mean squared error as loss function, a learning rate of 0.1, gamma set to 0, and with a max depth of 3 to avoid over-fitting on lexicon items/tokens occurring highly infrequently. SVM and KNN were run using the scikit-learn library. For SVM, we used the support vector regression (SVR) with an RBF kernel, the regularization parameter C set to 1, Epsilon parameter as 1, and no limit on maximum iterations. For k-Nearest Neighbors, the KNN regressor was employed using Euclidean distance, with the number of neighbors set to 5.

Standard deep learning CNN/RNN models, in concert with word embeddings, are used extensively for automated scoring of quality and/or traits in text \cite{alikaniotis2016automatic,taghipour-ng-2016-neural, dong2016automatic, berggren2019regression,yu2017deep, majumder2017deep}. The CNN \cite{yu2017deep} model used a static word embedding as input, run through a single 1D convolutional layer, followed by a max pooling and fully connected (dense) layers. The input static word embedding used was the Word2Vec word embedding with 100 dimensions. The CNN model was trained using a learning rate of 0.0005, a batch size of 300, and 20 epochs. The GRU model \cite{majumder2017deep} used the same input word embedding as the CNN, and featured a two layer bi-directional GRU model followed by a batch normalization layer and fully connected (dense) layers. 

For transformer-based, we incorporated BERT and RoBERTa \cite{devlin2018bert,liu2019roberta,yang2023getting,schmalz2021automatic}. Fine-tuned transformer models, as well as the pre-trained models' embeddings, and zero-shot transformer LLM assessments, have been used in an array of text scoring research \cite{yang2023getting,schmalz2021automatic,yancey2023rating}. Transfomer-based PLMs such as BERT and RoBERTa have attained state-of-the-art text sequence classification performance for an array of tasks including automated assessment of text quality and traits \cite{ schmalz2021automatic, uto2020neural, sung2019improving, dong2021automated, gupta2023data}. The two transformer PLMs were trained as follows. The BERT \cite{devlin2018bert} base model, with 110M parameters, and the RoBERTa  \cite{liu2019roberta} base model with 125M parameters were used. For both models, further fine-tuning on the AES training data was performed using 5-fold cross validation as done for the other models. Moreover, hyperparameters were tuned to balance under versus over-fitting by separating a portion of the training data into a separate validation set. Both models were run with learning rates set to 0.0005, and batch sizes of 300. In regards to the number of epochs, BERT was run for 30 epochs whereas RoBERTa was run for 20. The output of BERT and RoBERTa were 768-dimensional embeddings. We input these into a batch normalization layer and then a fully connected (dense) layer and output layer to get the prediction scores.

\subsection{Benchmarking}
Consistent with our first research question (and pursuant with our first research gap), the goal of the benchmark analysis depicted in the center of Figure \ref{fig:analysisFrame} was to assess the effectiveness of ML methods for AES on human-generated text. Consistent with prior work \cite{ijcai2019p879,Ramesh2022-kg}, two sets of metrics were employed. The first were error metrics: mean squared error (MSE) and mean absolute error (MAE). The second were agreement/correlational metrics: quadratic weighted kappa (QWK), Pearson correlation coefficient (PCC), and Spearman rank correlation (SRC).

\subsection{Statistical Analysis}
Consistent with RQ2, the goal of our statistical analysis was to parsimoniously examine the interplay between text genres, ML models used for assessment, and human versus GPT respondents (see center of Figure \ref{fig:analysisFrame}. We employ a mixed-effects model, taking Prompt Type, Prompt Respondent, and the ML model as the primary fixed effects and examining the possible two-way or three-way interaction between them. GPT generated text tend to differ from humans in response length on a given prompt, with GPT tending to be more verbose (i.e., lengthier). To account for this, the length of a generated text in words was included as a control variable in the ANOVA. We also controlled for the testbeds associated with each prompt (i.e., ASAP versus CLC-FCE). 

Overall, the factorial structure of the model was as follows:

 \begin{equation}
Y = A/S + A + B + C + D + E + A \times B + A \times C + B \times C + A \times B \times C + \epsilon
\label{eq:anova}
\end{equation}

Dependent variable \textit{Y} corresponds to the predicted score. \textit{S} stands for the 66 unique prompts, and \textit{A} stands for the prompt type. \textit{B} signifies the respondent type, while \textit{C} denotes the ML model. Considering the difference in answer length between human and ChatGPT generated text, we introduce text length, \textit{D}, as a control variable. Finally, we include testbed, \textit{E}, as a control variable to account for the two sources of prompts: AES and FCE. Additionally, the model includes a random intercept to account for the random effect of different prompts nested within each prompt type.

\subsection{Content Analysis}
The purpose of the content analysis is, as noted in RQ3, to compare and contrast the linguistic characteristics of GPT versus human-generated text. Our content analysis explores this question in two ways. First, we leverage the idea of parallel representations \cite{abbasi2010selecting,ahmad2020deep}. Guided by the intuition that although text is one dimensional, language is multi-dimensional -- spanning semantics (e.g., word senses, topics), sentiment, affect, psychological, pragmatic, syntactic, and stylistic elements -- parallel representations attempt to capture the richness of language in 1D text. 

Table \ref{tab:linguistic_representations} shows examples of parallel representations, where bolded tokens for parallel representations signify differences relative to the primary 1D word representation depicted in the first row of the table. As shown in the table, only some of the tokens in parallel representations convey new/different information, such as word sense, parts-of-speech, sentiment polarity, affect, and mapping to tags in different lexicons, including the linguistic inquiry word count (LIWC). In prior text classification work, univariate and multivariate measures of information or correlation have been used for feature selection of tokens \cite{abbasi2010selecting} for classifying sentiment, or to train embeddings \cite{ahmad2020deep} for detecting psychometrics. Here, we use the same idea, but with one important difference. Our binary class label is whether the text was authored by a human or GPT respondent. And we wish to use parallel representations to see how the expressive power of human generated text differs from that of GPT text, across the linguistic representations depicted in Table \ref{tab:linguistic_representations}. The idea being that higher expressive power connotes greater linguistic differences between human and GPT generated text.

Table~\ref{tab:linguistic_representations} delineates the comprehensive parallel representations generated for the sample input text, "It was so hot outside, it was like the Sahara desert. I got out of the car with a huge grin on my face." These representations encompass word-level tokens, named entities, hypernyms, and domain-specific lexicons. These elements are strategically employed to capture topical nuances at a macroscopic level, thereby facilitating the identification of underlying patterns and structures even in data-limited scenarios.

For example, named entities, extracted via Stanford CoreNLP~\cite{manning2014stanford}, amalgamate information at the granularity of persons, places, and organizations. Hypernyms, derived from WordNet's hierarchical structure~\cite{fellbaum2010wordnet}, enable the aggregation of entities based on "is-a" relationships. Domain lexicons, curated through rigorous data domain analysis~\cite{beijer2002hospitalisations}, enrich the semantic layer by linking terms like "face" to domain-specific labels such as "ANATOMY."

Crucially, the representations in Table~\ref{tab:linguistic_representations} maintain consistent lengths and are index-aligned. This alignment imbues them with a "parallel" nature, thereby facilitating the fusion and correlation of features. For instance, word tokens can be amalgamated with sense tags to construct a "Word \& Sense" representation, which leverages WordNet for enhanced word sense disambiguation. Similarly, a "Word \& NE" representation can be formulated by merging words with named entities, thereby offering multiple levels of semantic granularity.

Beyond these, sentiment and affective states serve as valuable supplementary representations. Sentiment analysis enables the quantification of user subjectivity and emotional tone, with words mapped to their corresponding sentiment scores as per SentiWordNet~\cite{baccianella2010sentiwordnet}. These scores can be categorized into high, medium, or low bins, culminating in a nuanced set of nine possible sentiment tags. For example, the term "hot" exhibits low negative polarity. Affective categories, derived from WordNet Affect~\cite{strapparava2004wordnet}, further enrich the representation, as evidenced by the mapping of the word "got" to the "SURPRISE" category in Table~\ref{tab:linguistic_representations}.

Syntactic representations incorporated include parts-of-speech (POS), Word \& POS, misspellings, and hapax legomena. Misspellings introduce sparsity and noise, particularly in user-generated content. To mitigate this, a spellchecking algorithm with support for word exclusions is employed. Given that the frequency of misspellings serves as a critical psychometric indicator, this representation is included. As illustrated in Table~\ref{tab:linguistic_representations}, the misspelled term "sahara" (needed to be capitalized) is corrected across all representations and tagged as a MISSPELLING. Hapax legomena are utilized to address sparsity arising from singleton words in the training set, and are tagged with a DIS label~\cite{popescu2008hapax}. The amalgamation of words with their corresponding POS tags serves as an additional disambiguation layer.

\begin{table}[t!]
\centering
\footnotesize
\begin{tabular}{|p{2.5cm}|p{13cm}|}
\hline
\textbf{Representation} & \textbf{Example} \\
\hline
Word & it was so hot outside , it was like the sahara desert . i got out of the car with a huge grin on my face . \\
\hline
Word \& Sense & it was so \textbf{hot\_\_03 outside\_\_09} , it was \textbf{like\_\_02} the \textbf{sahara\_\_01 desert\_\_01} . i \textbf{got\_\_01} out of the \textbf{car\_\_01} with a \textbf{huge\_\_01 grin\_\_01} on my \textbf{face\_\_04} . \\
\hline
Word \& POS & \textbf{it\_\_PRON was\_\_AUX so\_\_ADV hot\_\_ADJ outside\_\_ADV ,\_\_PUNCT it\_\_PRON was\_\_AUX like\_\_ADP the\_\_DET sahara\_\_PROPN desert\_\_NOUN .\_\_PUNCT i\_\_PRON got\_\_VERB out\_\_ADP of\_\_ADP the\_\_DET car\_\_NOUN with\_\_ADP a\_\_DET huge\_\_ADJ grin\_\_NOUN on\_\_ADP my\_\_PRON face\_\_NOUN .\_\_PUNCT} \\
\hline
Word \& NE & it was so hot outside , it was like the \textbf{sahara|\_|\_LOC} desert . i got out of the car with a huge grin on my face . \\
\hline
Hypernym & it was so hot outside , it was DESIRE the sahara \textbf{BIOME} . i \textbf{CHANGE\_STATE} out of the \textbf{MOTOR\_VEHICLE} with a huge \textbf{FACIAL\_EXPRESSION} on my \textbf{SURFACE} . \\
\hline
Named Entities & it was so hot outside , it was like the \textbf{LOC} desert . i got out of the car with a huge grin on my face . \\
\hline
Domain Lexicons & it was so hot outside , it was like the sahara desert . i got out of the car with a huge grin on my \textbf{ANATOMY} . \\
\hline
Sentiment & it was so \textbf{LPOSLNEG LPOSLNEG} , it was \textbf{LPOSLNEG} the \textbf{LPOSLNEG LPOSLNEG} . i \textbf{LPOSLNEG} out of the \textbf{LPOSLNEG} with a \textbf{LPOSLNEG LPOSLNEG} on my \textbf{LPOSLNEG} . \\
\hline
Affect & it was so hot outside , it was like the sahara desert . i \textbf{SURPRISE} out of the car with a huge grin on my face . \\
\hline
POS & \textbf{PRON AUX ADV ADJ ADV PUNCT PRON AUX ADP DET PROPN NOUN PUNCT PRON VERB ADP ADP DET NOUN ADP DET ADJ NOUN ADP PRON NOUN PUNCT} \\
\hline
MISSPELLING & it was so hot outside , it was like the \textbf{MISSPELLING} desert . i got out of the car with a huge grin on my face . \\
\hline
Legomena & it was so hot outside , it was like the \textbf{DIS} desert . i got out of the car with a huge grin on my face . \\
\hline
\end{tabular}
\caption{Illustration of select parallel representations for an example sentence}
\label{tab:linguistic_representations}
\end{table}

To utilize the representations, we calculate a  weight \cite{ahmad2020deep} for each token in each representation. After obtaining this weight for each token, we use a threshold to filter out the tokens providing limited additional information (i.e., those below the threshold). The expressive power of that representation for differentiating human and GPT text is calculated as the ratio of tokens above the threshold relative to the total number of tokens.

More formally, we quantify the linguistic expressive power of text, using parallel representations as follows. As a first step, the weight of every token across parallel representations is calculated. Given the set of $m$ representations $R = \{r_1, r_2, ... r_m\}$, where $r_j$ denotes any parallel representation, we extract all tokens (i.e., 1-gram features). Any element $f_{ij} \in r_j$ represents the $i^{th}$ unigram/token feature for representation $r_j$. The initial weight of $f_{ij}$ is calculated as \cite{ahmad2020deep}:

\begin{equation}
w\left(f_{i j}\right)=\max _{c_a, c_b}\left(p\left(f_{i j} \mid c_a\right) \log \left(\frac{p\left(f_{i j} \mid c_a\right)}{p\left(f_{i j} \mid c_b\right)}\right)\right)+s\left(f_{i j}\right),
\end{equation}
where function $s$ is the mean semantic orientation score across all token-senses $v$, computed as the difference between the positive and negative polarity scores for sense $q$ of token $f_{ij}$ in SentiWordNet, $c_a$ and $c_b$ are among the set of $C$ class labels, $c_a \neq c_b$.

\begin{equation}
    s(f_{ij}) = \sum_{q}^{v} \frac{pos(f_{ij},q) - neg(f_{ij},q)}{dv}
\end{equation}
The first part of the weighting equation considers the discriminatory potential of the feature based on its log-likelihood ratio, whereas the second part factors in the semantic orientation to ensure that features with opposing orientation (e.g., "like" versus "don't like") are differentiated in terms of overall weights. 

Next, in order to capture the unique information for parallel representations beyond word, we adjust their weights accordingly. Let us assume that for any representation $r_j$ in $R$, $j=1$ denotes the word representation and $j>1$ signifies other parallel representations. For each feature token $f_{ij}$ in $j>1$, we compute:

\begin{equation}
    p\left(f_{i j}\right)= 
\begin{cases}
    1,& \text{if } w\left(f_{i j}\right) > t_w  \wedge \max _{f_{uv}}\left ( \rho(f_{ij},f_{uv})  \right ) \le t_c   \\
    0,              & \text{otherwise}
\end{cases}
\end{equation}

where $t_w$ and $t_c$ are pre-defined thresholds for weight and correlation, respectively, $\rho$ is the Pearson's correlation coefficient, $u$ is any token in representation $v$, and $j \neq v$.

Finally, we define the expressive power of any representation $r_j$ for $j>1$ as:

\begin{equation}
    e\left(r_j\right)= \sum_{i}\frac{p\left(f_{i j}\right)z_{i j}}{\sum_{i}z_{i j}}
\end{equation}

where $z_{i j}$ are the total number of occurrences of $f_{ij}$ in the data corpus, and hence, each $e\left(r_j\right)$ is a value between 0 and 1 denoting the proportion of unique token occurrences of that representation in the corpus that contribute additional information atop the baseline word representation. Larger values denote greater differences in linguistic characteristics across class labels $c_a$ versus $c_b$. In our subsequent content analysis, we assess the expressive power of GPT versus human text, while including differences between other human demographic classes such as age (older versus younger authors) and gender (male versus female authors) as reference groups for comparison. 

As an additional content analysis, we leverage advancements in Natural Language Processing to visualize and analyze attention mechanisms in transformer-based models \cite{niu2021review}. Specifically, we extend the work of bertviz\cite{vig2019multiscale} to compare parallel token-based representation weights, denoted as \( w(f_{ij}) \), against the token attention weights generated by the transformer's attention layers. We employ bertViz's multiple views, such as the Attention-Head View and Model View, to visualize these comparisons. While bertViz designed to visualize single attention-head view, which can add understanding of how the specific tokens are processing by transformer-based models, it may lack of the capability of reflect the overall attention received by the specific tokens. Hence, we develop equation ~\ref{equ: attention_score} to provide each specific token an attention score.

Let \( A_i \) represent the aggregated attention score for token \( i \), where \( i \) is also present in the parallel representation \( r_1 \) with \( w(f_{ij}) > t_w \), the token weight threshold. The aggregated attention \( A_i \) for a token \( i \) is defined as:

\begin{equation}
A_i = \frac{1}{N} \sum_{l=1}^{N} \text{mean}(l_{ik})
\label{equ: attention_score}
\end{equation}

Where \( A_i \) is the aggregated attention score for token \( i \), \( N \) is the total number of attention layers in the transformer model, \( l_{ik} \) denotes the attention scores for token \( i \) at position \( k \) in layer \( l \), and \( \text{mean}(l_{ik}) \) computes the average attention score across all attention heads for token \( i \) at position \( k \) in layer \( l \).

We compute \( A_i \) for each token of interest and store the results in a dictionary that maps each token to its respective list of aggregated attention scores. This allows us to contrast the assessments made by the transformer model for human-generated text versus text generated by GPT models.




\section{Results - Benchmark Evaluation and Empirical Analysis}
\subsection {Benchmark Evaluation Results}
Our first research question asks about the effectiveness of state-of-the-art feature-based and deep learning methods for AES. As noted in Section 3.3, the seven ML models were each trained separately on the ASAP and CLC-FCE testbeds using 5-fold cross-validation. This was done because training each classifier within a given testbed offered better performance vis-à-vis training them across a consolidated training set encompassing both testbeds. Additionally, the dependent variable within both testbeds was standardized to a 0-1 continuous scale. As noted in Section 3.4, we employed two error metrics, MSE and MAE, and three agreement/correlation measures (QWK, PCC, SRC) \cite{ijcai2019p879,Ramesh2022-kg}. Given MSE and MAE are error metrics, values closer to 0 denote better performance. For QWK, PCC, and SRC, values closer to 1 indicate better performance, whereas values closer to 0 signify random performance. 

The results appear in Table \ref{table:bench_results}. For both data sets, across all five metrics, the two transformer PLM-based models, BERT and RoBERTa, attained the best performance. Their PRC and SRC values in the 0.5 to 0.76 are on par with best-in-class text scoring results attained on problems such as psychometric NLP and personality detection \cite {lalor2022benchmarking}. Additionally, the QWK results for BERT and RoBERTa are comparable to the best ASAP full dataset results attained in prior studies \cite{ijcai2019p879,Ramesh2022-kg}.However, it is worht noting that direct comparisons are difficult to make because prior studies have used different problem formulations, training-testing splits, and different testbeds (e.g., even within ASAP) \cite{Ramesh2022-kg}. This lack of prior streamlined large-scale benchmarking was/is the impetus for our RQ1. 

In regards to other ML models, the CNN/RNN methods incorporated, namely CNN and GRU, outperformed feature-based methods such as SVR, XGB, and KNN. This is also aligned with prior surveys of AES \cite{ijcai2019p879,Ramesh2022-kg} and related areas such as personality \cite{yang2023getting}. Interestingly, for all methods on all five metrics, results were somewhat better on the ASAP testbed as compared to CLC-FCE. This could be due to the greater abundance of available training data. In prior AES studies, ASAP has also seen greater usage \cite{uto2021review,ijcai2019p879,Ramesh2022-kg}. Overall, the benchmarking results lend credence to the ML assessment models incorporated in our study. The results suggest that it is reasonable to include  ML assessment scores as part of our statistical analysis model (see ensuing section for details). 

\begin{table}[]
\small
\centering
\begin{tabular}{|lllllll|}
\hline
\multicolumn{7}{|c|}{ASAP Testbed}      \\
   & \textbf{Models}  & \textbf{MSE}  & \textbf{MAE}  & \textbf{QWK}  & \textbf{PCC}  & \textbf{SRC}  \\ \hline
\multirow{2}{*}{Transformer PLM} & BERT & \textbf{0.0241} & \textbf{0.1187} & 0.7034  & 0.7648  & \textbf{0.7685} \\
   & RoBERTa  & 0.0252  & 0.1236  & \textbf{0.7105} & \textbf{0.7663} & 0.7612  \\ \hline
\multirow{2}{*}{Deep Learning CNN/RNN} & CNN  & 0.0280  & 0.1299  & 0.6856  & 0.7214  & 0.7158  \\
   & GRU  & 0.0274  & 0.1306  & 0.6844  & 0.7351  & 0.7299  \\ \hline
\multirow{3}{*}{Feature-based ML}  & SVR  & 0.0293  & 0.1325  & 0.6071  & 0.7143  & 0.7081  \\
   & XGB-RF-Regressor & 0.0329  & 0.1396  & 0.5748  & 0.6575  & 0.6553  \\
   & KNN-Regressor & 0.0372  & 0.1491  & 0.6168  & 0.7000  & 0.6914  \\ \hline
\multicolumn{7}{|c|}{FCE Testbed}     \\
   & \textbf{Models}  & \textbf{MSE}  & \textbf{MAE}  & \textbf{QWK}  & \textbf{PCC}  & \textbf{SRC}  \\ \hline
\multirow{2}{*}{Transformer PLM} & BERT & 0.0277 & 0.1350 & 0.3123  & 0.5152  & \textbf{0.5227} \\
   & RoBERTa  & \textbf{0.0266} & \textbf{0.1304} & \textbf{0.3307} & \textbf{0.5158} & 0.5214  \\ \hline
\multirow{2}{*}{Deep Learning CNN/RNN} & CNN  & 0.0381  & 0.1563  & 0.0956  & 0.1313  & 0.1133  \\
   & GRU  & 0.0377  & 0.1556  & 0.0945  & 0.1223  & 0.1195  \\ \hline
\multirow{3}{*}{Feature-based ML}  & SVR  & 0.0339  & 0.1553  & 0.0475  & 0.3212  & 0.3226  \\
   & XGB-RF-Regressor & 0.0344  & 0.1480  & 0.0943  & 0.1691  & 0.1983  \\
   & KNN-Regressor & 0.0356  & 0.1512  & 0.1490  & 0.2067  & 0.2034  \\ \hline
\end{tabular}
\caption{Benchmarking Results Comparing Performance of Different ML Methods on Human Generated Text}
\label{table:bench_results}
\end{table}

\begin{figure}[ht]
        \centering
        \includegraphics[width=\textwidth]{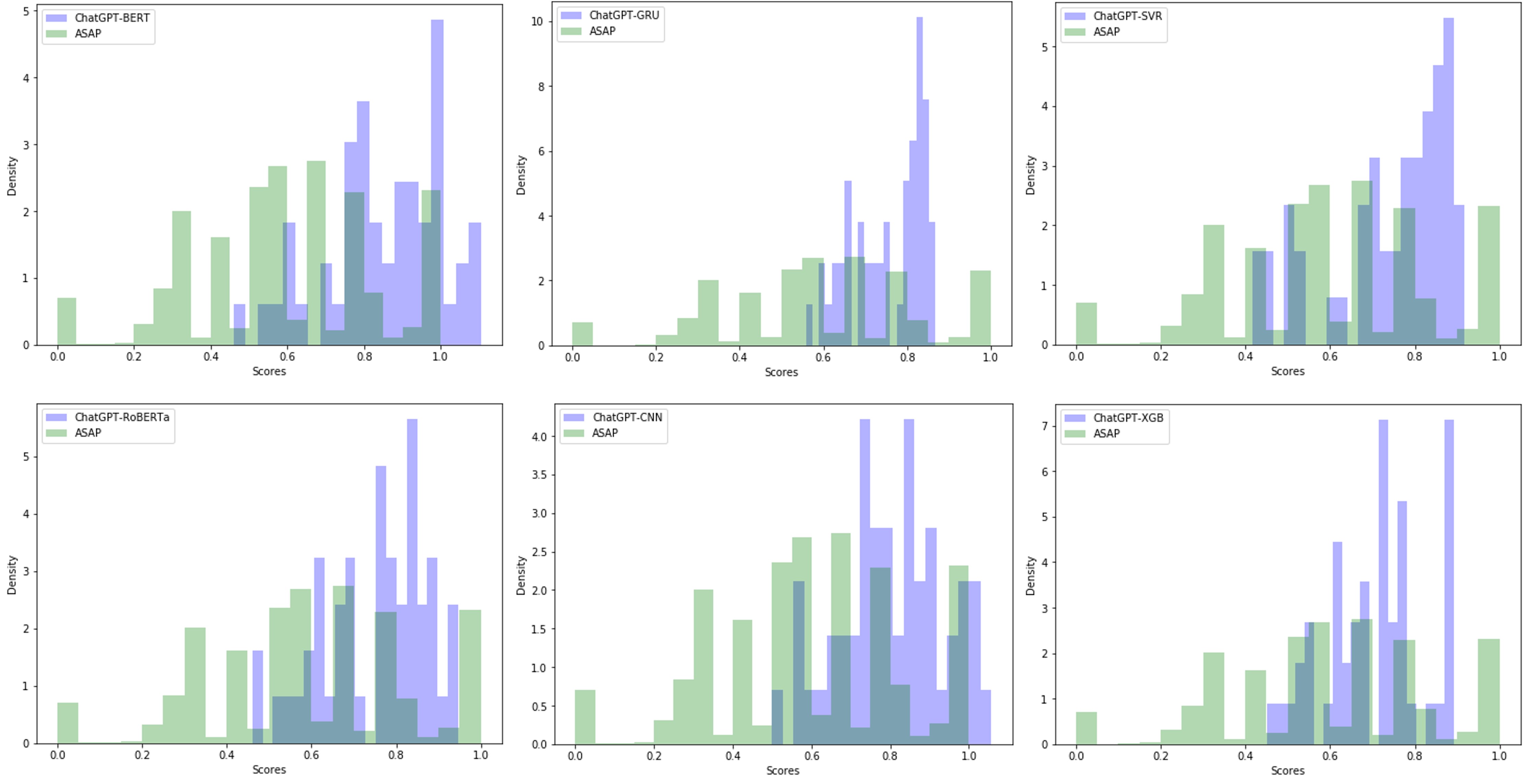}
        \caption{ML Assessments of GPT Generated Text Versus Expert Ratings of Human Text}
        \label{fig:researchBars}
\end{figure}

Figure \ref{fig:researchBars} shows the distribution of human text gold standard scores, scaled to a 0-1 range (green background bars), on the ASAP testbed. Also depicted are the ChatGPT generated texts’ ML model scores for BERT and RoBERTa (first column top and bottom charts, respectively), GRU and CNN (middle column), and SVR and XGB. Note that because the human text scores are the gold standard, they do not vary by model whereas the ChatGPT scores do. Looking at the distributions, we can see that the ML assessment models typically rate the ChatGPT text higher. Further, the ML assessments of ChatGPT text also follow a tighter distribution with a smaller range and less variance. 

\subsection{Results - Empirical Analysis}

\noindent Our second research question asked how AES models trained on human-generated text rated content generated by GPT models, and how the effects were moderated by document genres and different categories of ML models used for assessment. As noted in Section 3.4, and presented in Figure \ref{fig:analysisFrame}, our analysis cube spans three dimensions: prompt genres, ML assessment models, and human versus ChatGPT generated text. Model-free analysis methods are ill-equipped for parsimoniously considering the interactions between these three dimensions, as well as the impact of other control variables such as the type of testbed (ASAP versus CLC-FCE) and other considerations (e.g., essay length). Accordingly, we used a 3-way split-plot ANOVA design to assess variations in ML assessment model scores given the impacts of the prompt type, prompt respondent (text author - GPT or human), and the ML model used to perform the scoring assessment. The input for the ANOVA model were an ML-text score tuple. Given we had 7 ML models for scoring (see rows in Table \ref{table:bench_results}, this meant 7 rows per 13,847 total documents in our human plus GPT testbeds when using ChatGPT powered by GPT-3.5, and 15,333 total documents when using ChatGPT powered by GPT-4 in additon to GPT-3.5. Similar to our benchmark analysis, ML model scores were standardized to a range between 0 and 1. The 66 different prompt IDs were a between factor that was nested under the 6 prompt genres mentioned above. The prompt respondent and ML model used were within factors, meaning that each response generated either from GPT or human was repeatedly scored using different ML models pre-trained with human responses. In the two following sub-sections, we report the 3-way split-plot ANOVA results when using GPT-3.5 versus humans, and as a robustness check, also when using GPT-4 in addition to GPT-3.5.

\subsubsection{Results when Comparing Human and GPT-3.5 Generated Text}

The overall significance results for main effects, two-way, and three-way interactions appear in Table \ref{table:anova_results}. As noted in Section 3.4 when discussing \eqref{eq:anova}, \textit{A,B,C} correspond to the three dimensions of our analysis cube, namely prompt type, respondent type, the ML assessment model, respectively. \textit{D,E} relate to the text length and the testbed source for the prompts (i.e., ASAP or CLC-FCE). For precision, here we only present the ANOVA results in Table \ref{table:anova_results}. Effect size plots appear in figures discussed subsequently. The table depicts statistical significances for the main effects as well as two-way and three-way interactions pertaining to \textit{A,B} and \textit{C}. Overall, with the exception of the prompt type main effect, all other factors were significant (i.e., p-values < 0.05).

\begin{table}[ht]
\centering
\begin{tabular}{lrrrrrr}
\hline
 \textbf{Source} & \textbf{SS} & \textbf{MS} & \textbf{NumDF} & \textbf{DenDF} & \textbf{F value} & \textbf{Pr($>$F)} \\ 
\hline
A (prompt type)  & 0.31 & 0.06 & 5 & 2 & 3.8566 & 0.2186 \\ 
B (human/GPT text)  & 1.76 & 1.76 & 1 & 82419 & 109.7398 & $<$2e-16*** \\ 
C (scoring model)  & 1.82 & 0.36 & 5 & 97458 & 22.7752 & $<$2e-16*** \\ 
D  & 515.82 & 515.82 & 1 & 97309 & 32200.2932 & $<$2e-16*** \\ 
E  & 1.72 & 1.72 & 1 & 41 & 107.6747 & 4.29e-13*** \\ 
A$\times$B & 14.92 & 2.98 & 5 & 63321 & 186.2416 & $<$2e-16*** \\ 
A$\times$C & 6.37 & 0.25 & 25 & 97458 & 15.8982 & $<$2e-16*** \\ 
B$\times$C & 13.66 & 2.73 & 5 & 97458 & 170.5444 & $<$2e-16*** \\ 
A$\times$B$\times$C& 5.68 & 0.23 & 25 & 97458 & 14.1775 & $<$2e-16*** \\ 
\hline
\end{tabular}
\caption{Results for Statistical Model with Human Versus GPT-3.5 Generated Text}
\label{table:anova_results}
\end{table}

\begin{figure}[ht]
        \centering
        \includegraphics[width=7cm]{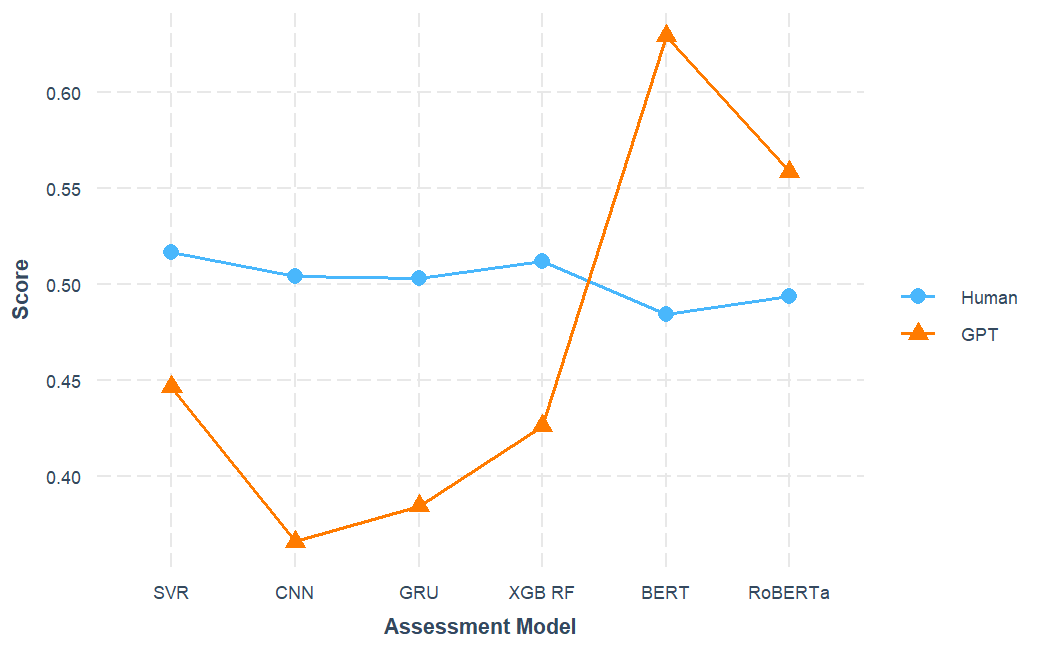}
        \caption{ML Model Comparison By Respondent Type}
        \label{fig:modelgpt}
\end{figure}

\begin{figure}[ht]
\centering
\includegraphics[width=7cm]{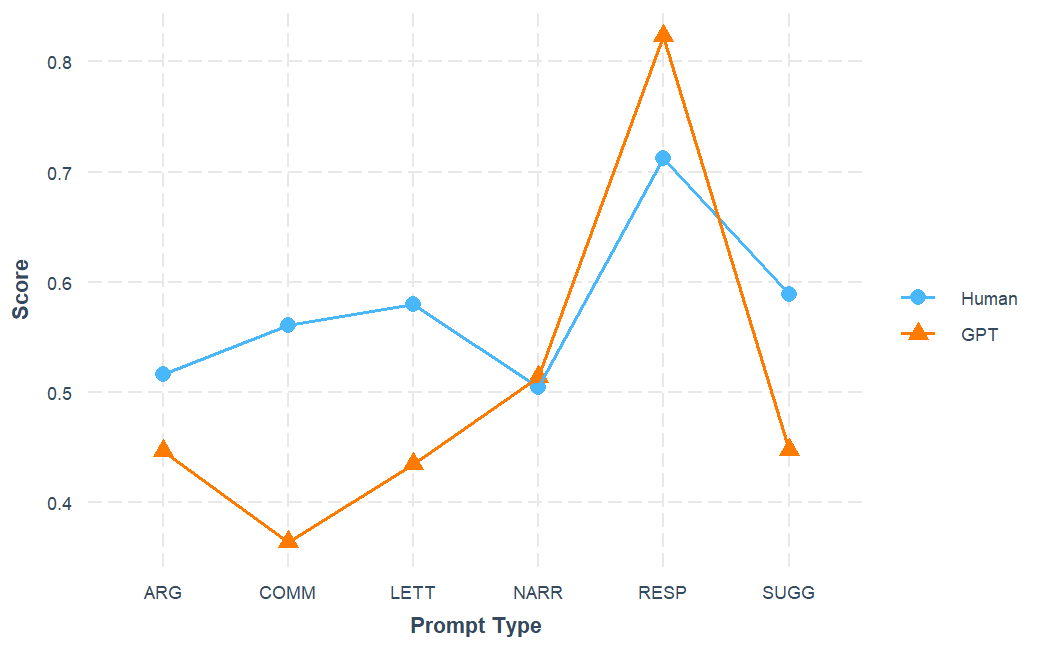}
\caption{Interaction Effect Between Respondent Types and Prompt Types}
\label{fig:typegpt}
\end{figure}

\begin{figure}[ht]
\centering
\includegraphics[width=7cm]{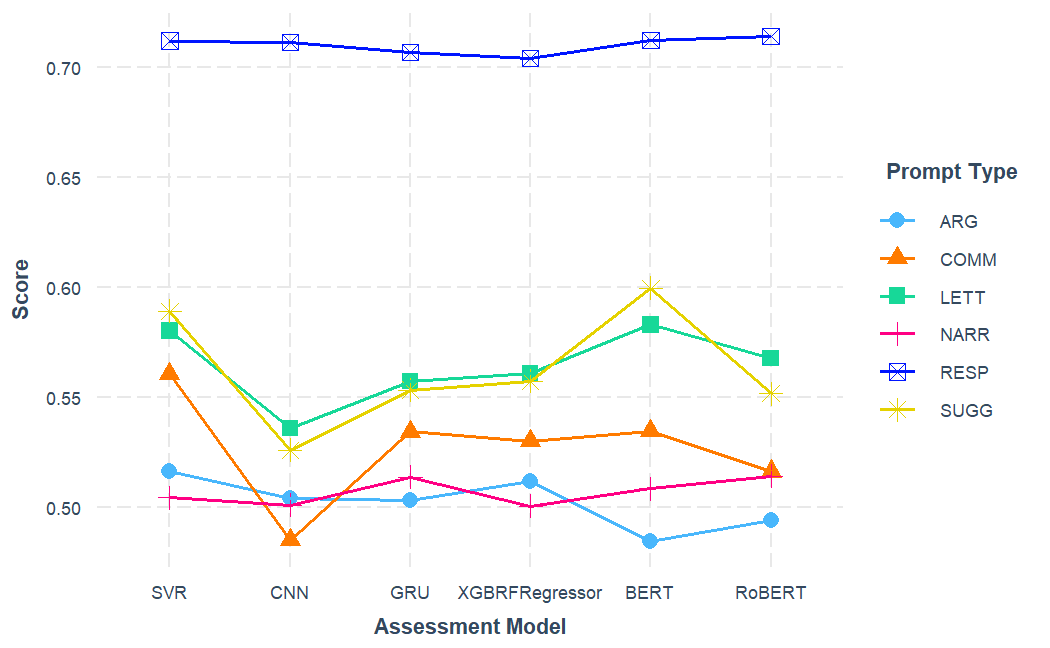}
\caption{Interaction Effect Between ML Models and Prompt Types}
\label{fig:typemodel}
\end{figure}

Figures \ref{fig:modelgpt}, \ref{fig:typegpt}, and \ref{fig:typemodel} show effect sizes for the three two-way figures. Figure \ref{fig:modelgpt} shows results for the ML assessment model (x-axis) comparing the human and GPT text responses. That is, the \textit{B}x\textit{C} interaction. The models along the x-axis are intentionally grouped by type, with feature-based being left-most, followed by CNN/GRU in the middle, and the two transformer models on the right. For brevity, kNN was excluded from the plots. The results show that feature-based ML assessment models such as SVR and XGB tend to score human-generated text about 10-15\% higher than GPT generated responses (about 7-10 points higher; for instance XGB scores human texts at 0.72 and GPT ones at around 0.65). This disparity becomes even more pronounced when using CNN and GRU to perform the assessment. On average, these deep learning models score human responses about 26-32\% higher (i.e., about 15-25 points higher). These results seem to suggest that perhaps due to the out of training sample nature of GPT generated text responses, the feature and ML models tend to rate these text lower. However, when looking at the transformer PLM results, both BERT and RoBERTa tend to score the GPT essays markedly higher than the human ones. For BERT, the difference is approximately 10 points (0.78 versus 0.68) or about 15\%. This “flip” relative to the feature-based and CNN/RNN deep learning models is important because as noted in our benchmark analysis, the transformer PLM models attain state-of-the-art performance. 

This “affinity” for GPT generated text, relative to human responses, as depicted in Figure \ref{fig:modelgpt}, could be attributable to a few possible reasons. First, the feature-based ML models and CNN/GRU deep learning models have pre-defined features/vocabularies based on their input feature sets and/or static word embeddings. Conversely, transformer PLMs use a much larger pre-training corpora and also rely on word piece to avoid out-of-vocabulary (OOV) concerns \cite{devlin2018bert,liu2019roberta}. Hence, the absence of GPT generated text in the training phase may be less of a concern for transformer PLMs tasked with assessing such text. Second, as noted in our review of related work, BERT and RoBERTa share a fair amount of common pre-training data with GPT models, namely the Wikipedia corpus, 11K Book and/or Book 1 and Book 2 corpora, and common crawl and web text data \cite{radford2019language,brown2020language}. Admittedly, with GPT-3.5 onwards, the true extent of the training corpora for ChatGPT models is unclear, and the proportion of overlap is likely smaller. Nevertheless, familiarity in pre-training sources, as well as relative commonalities in the underlying attention/learning mechanisms (at least vis-à-vis CNN/GRU and SVR/XGB), may contribute to higher quality score assessments for GPT generated text when evaluated by BERT/RoBERTa. Third, it could be that because the transformer PLMs are more accurate (as per results in Table \ref{table:bench_results}), they are less prone to over-fitting, and hence, are more capable of generalizing to the out-of-sample GPT text and assessing its quality. We explore the first two possible explanations in greater detail in the ensuing section related to RQ3.

Figure \ref{fig:typegpt} shows the results for human and GPT-generated text across the six genres of prompts. This corresponds to the \textit{A}x\textit{B} factor row in Table \ref{table:anova_results}. For most genres, including argumentative (ARG), commentary (COMM), letter (LETT), and suggestion (SUGG) writing, on average, human texts scored 7 to 17 points (10-20\%) higher. For narrative writings (NARR), results for humans and GPT were comparable. In contrast, GPT text scored nearly 10 points higher than human generated content for response (RESP) writing. Figure \ref{fig:typemodel} shows the \textit{A}x\textit{C} interaction. The x-axis depicts ML models and the different series show prompt genres. In terms of impact of models on genre-level performance, many of the lines are reasonably flat, signifying consistent results across different models. Exceptions include the CNN model rating letter, suggestion, and comment texts lower, and BERT scoring letter and suggestion relatively higher. In regards to prompt genre trends, response (RESP) consistently attained the highest model predicted scores. It is worth noting that response was the only prompt type genre in our testbed that did not appear in CLC-FCE. Hence, the generally higher scores may also be a function of reliance on text from a single testbed. On the other end of the spectrum, response genre text was scored lower by all models.   


\begin{figure}[ht]
\centering
\includegraphics[width=10cm]{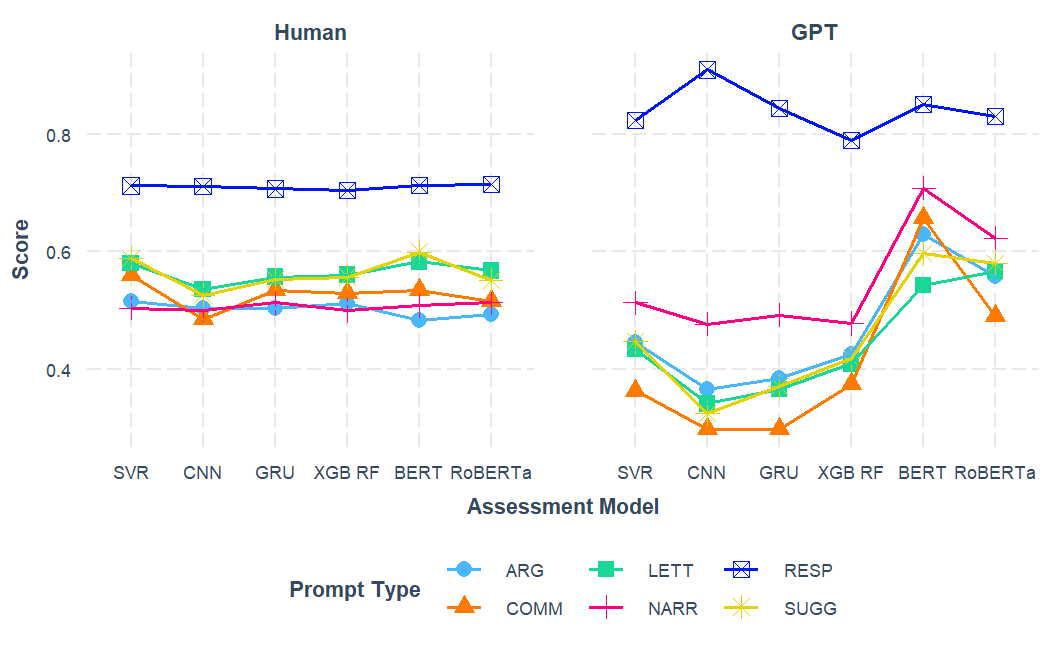}
\caption{Interaction Effect Between ML Models, Respondent Types and Prompt Types}
\label{fig:threeway}
\end{figure}

Figure \ref{fig:threeway} shows the three-way interactions. The left panel shows models (x-axis) and prompt genres (different series lines) for human-generated text, whereas the right panel depicts the same for GPT-generated text. Hence, collectively, the figure shows the \textit{A}x\textit{B}x\textit{C} row from Table \ref{table:anova_results}. Results in the left panel of Figure \ref{fig:threeway} are similar to the overall two-way \textit{A}x\textit{C} interaction results depicted in Figure \ref{fig:typemodel}. However, interestingly, the SVR/XGB/CNN/GRU models score GPT text for certain genres considerably lower than human text: argumentative, letter, comment, and suggestion. However, on all four of these aforementioned genres, as well as narratives, the two transformer-based PLM models score the GPT text relatively higher compared to the feature-based and CNN/GRU deep learning models. In general, BERT and RoBERTa scored GPT text considerably higher than human-generated content on response, narrative, and argumentative genres. For instance, on narratives, GPT text was scored about 15 points higher. On response texts, it was about 10 points higher. BERT and RoBERTa also scored GPT text fairly close to human text on the other genres. Interestingly, all models rated human and GPT response text the highest (see top trend line in both panels in Figure \ref{fig:threeway}). However, GPT text for the response genre was unanimously scored higher than human text across all ML models. Overall, the three-way interaction results shed further light on how and when ML assessment models deem GPT generated text quality to be better than that of human text, and how these effects are moderated by the model type and prompt genres. 

\subsubsection{Results when Comparing Human, GPT-3.5, and GPT-4 Generated Text}

As a robustness check, we wanted to see if our results for human versus GPT were consistent if using ChatGPT powered by GPT-4 versus if/and using GPT-3.5. Hence, we reran our ANOVA models with one notable difference; for respondent type, we included human, GPT-3.5, and GPT-4 as the three categories of text generators. The overall significance results for main effects, two-way, and three-way interactions appear in Table \ref{table:anova_results_gpt4}. Once again, \textit{A,B,C} correspond to the three dimensions of our analysis cube, namely prompt type, respondent type, the ML assessment model, respectively. \textit{D,E} relate to the text length and the testbed source for the prompts (i.e., ASAP or CLC-FCE). The table depicts statistical significances for the main effects as well as two-way and three-way interactions pertaining to \textit{A,B} and \textit{C}. Overall, consistent with Table \ref{table:anova_results} appearing earlier, with the exception of the prompt type main effect, all other factors were significant (i.e., p-values < 0.05).

\begin{table}[ht]
\centering
\begin{tabular}{lrrrrrr}
\hline
 \textbf{Source} & \textbf{SS} & \textbf{MS} & \textbf{NumDF} & \textbf{DenDF} & \textbf{F value} & \textbf{Pr($>$F)} \\ 
\hline
A  & 0.27 & 0.05 & 5 & 2 & 3.3867 & 0.2435 \\ 
B  & 1.03 & 0.52 & 2 & 101104 & 32.6008 & 7.018e-15*** \\ 
C  & 60.80 & 12.16 & 5 & 102317 & 766.2843 & $<$2e-16*** \\ 
D  & 488.21 & 488.21 & 1 & 102189 & 30766.2560 & $<$2e-16*** \\ 
E  & 1.01 & 1.01 & 1 & 41 & 63.5332 & 7.107e-10*** \\ 
A $\times$ B & 31.01 & 3.10 & 10 & 98035 & 195.4344 & $<$2e-16*** \\ 
A $\times$ C & 8.34 & 0.33 & 25 & 102317 & 21.0190 & $<$2e-16*** \\ 
B $\times$ C & 48.54 & 4.85 & 10 & 102317 & 305.9125 & $<$2e-16*** \\ 
A $\times$ B $\times$ C& 11.78 & 0.24 & 50 & 102317 & 14.8514 & $<$2e-16*** \\ 
\hline
\end{tabular}
\caption{Results for Statistical Model with Both GPT-3.5 and GPT-4 Included}
\label{table:anova_results_gpt4}
\end{table}

\begin{figure}[ht]
\centering
\includegraphics[width=10cm]{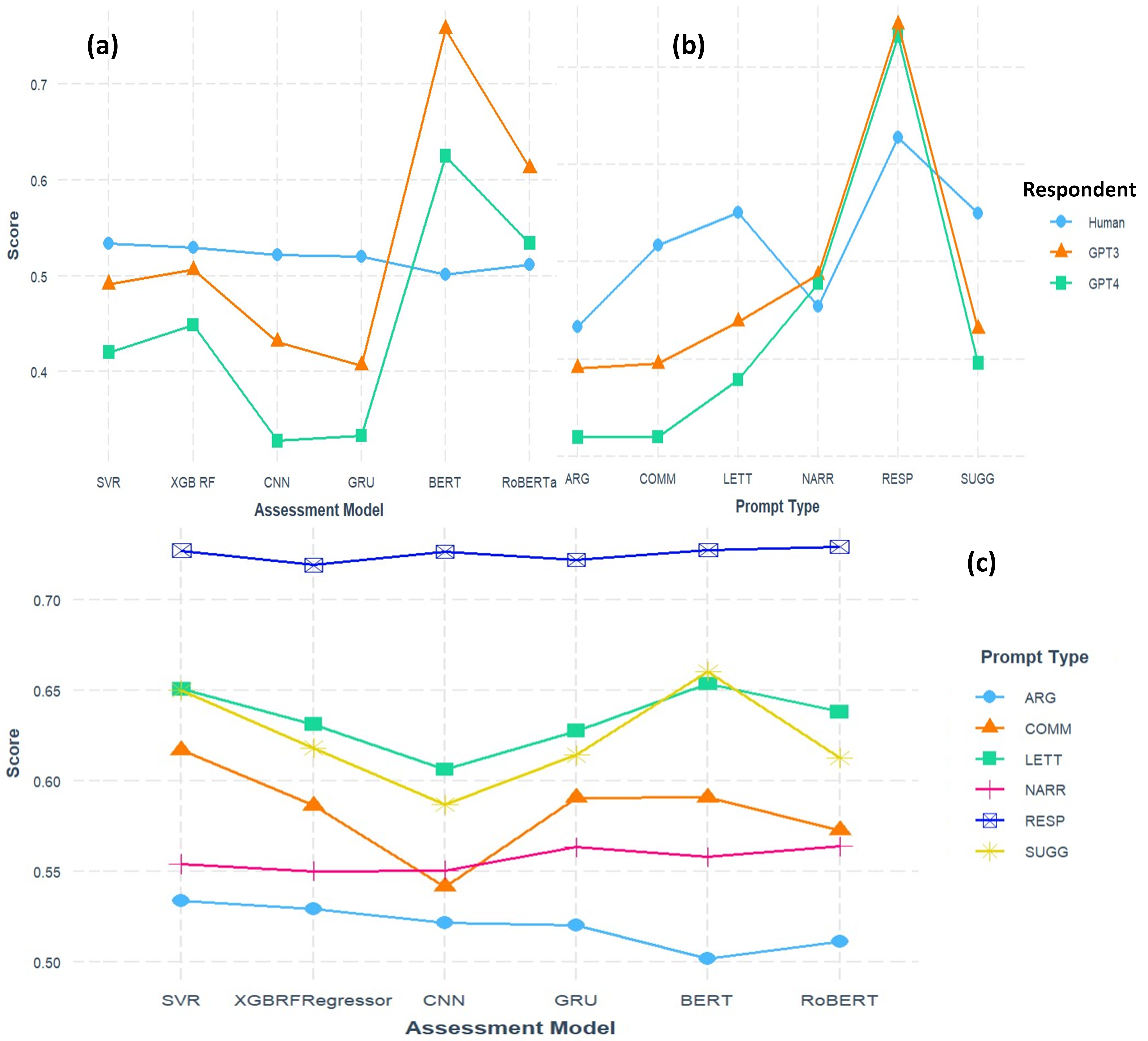}
\caption{Two-way Interaction Effect Plots when GPT-4 is Included}
\label{fig:twowayGPT4}
\end{figure}

Figure \ref{fig:twowayGPT4} shows the three two-way interaction plots for respondent-assessment (a), respondent-genres (b), and assessment-genres (c). The top two charts depict three series due to the inclusion of the three respondent categories: human, GPT-3.5, and GPT-4. Looking at the two-way interaction plots between assessment model and respondent type, depicted in chart (a), we observe a similar pattern to that depicted earlier. Feature and CNN/RNN-based assessment models score human-generated text higher than essays produced by both GPTs, whereas the the transformer-based language models (BERT and RoBERTa) rate the GPT generated text higher despite the GPT text being outside the training data used to fine-tune these two transformer-based assessment models. Interestingly, this effect is less pronounced for GPT-4 text as compared to essays generated by GPT-3.5. This could be due to greater differences in the underlying training data used by the BERT/RoBERTa models in comparison with GPT-4, relative to GPT-3.5, or differences in how ChatGPT using GPT-4 was trained/tuned relative to the version powered by GPT-3.5. It would be interesting to see if this trend continues with future work. Whether or how larger assessment models are more akin to GPT-3.5 or GPT-4 may affect quality scores for human versus LLM-generated text. 

Looking at charts (b) and (c) in Figure \ref{fig:twowayGPT4}, the overall respondent-genre and assessment-genre effects are similar to those observed earlier in Section 4.2.1. For instance, in regards to respondent-genre, chart (b), the argumentative (ARG), commentary (COMM), letter (LETT), and suggestion (SUGG) genres of prompt types were scored higher for humans versus GPTs, the respondent groups had comparable scores on narrative (NARR) writing, and the two GPT models scored markedly higher on response (RESP) writing. Similarly, in regards to the assessment-genre interaction effects, consistent with earlier results, response (RESP) writing attained the highest scores across models, whereas argumentative (ARG) text yielded the lowest scores.

\begin{figure}[ht]
\centering
\includegraphics[width=13.5cm]{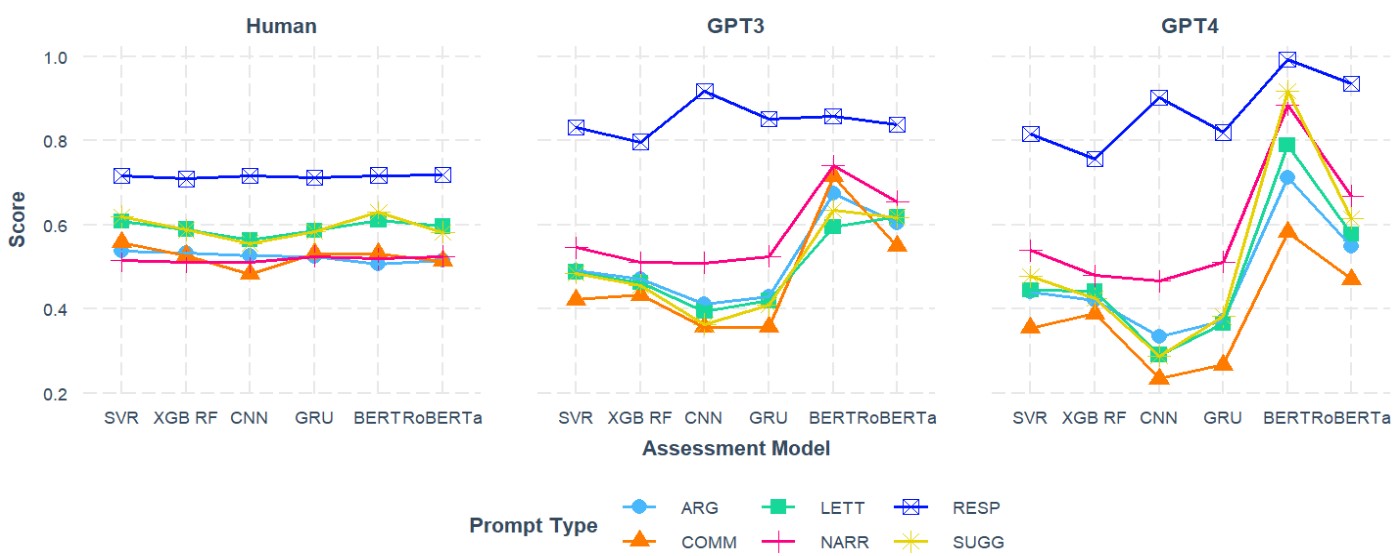}
\caption{Interaction Effect Between ML Models, Respondent Types (including GPT-4) and Prompt Types}
\label{fig:threewayGPT4}
\end{figure}

Finally, Figure \ref{fig:threewayGPT4} shows the three-way interaction plots for our three dimensional analysis cube when including the three respondent types. Relative to the GPT-3.5 text, the GPT-4 generated essays score higher when assessed by BERT and RoBERTa for response prompts (RESP). BERT also scores the GPT-4 text higher for narrative, suggestion, and letter writing. For argumentative and communication genres, BERT scored text from GPT-3.5 higher than that generated by GPT-4. In the next section, we use feature importance and attention analysis methods to delve deeper into linguistic differences between human and GPT text, and why the transformer model may be scoring GPT text higher.  

\section{Results - Content Analysis}
Our third research question asked what linguistic categories are most different for human versus GPT generated text. As discussed in section 3.6, our content analysis involved two parts. In the first, we computed the expressive power of parallel representations. In the second, we compared the parallel representation weights with the attention weights from the BERT assessment model.

\subsection{Content Analysis Using Parallel Representations}
The purpose of the parallel representation-based analysis of expressive power was to see how linguistically similar or different human versus GPT generated texts are, and across which linguistic categories. The idea being that if there are more features that can discern the respondent type (i.e., higher expressive power), this signifies greater differences between the two groups. We set our two class labels as human and GPT generated text, respectively. We used 1,537 essays generated by GPT-3.5 and 15,437 human essays from the FCE and ASAP corpora. In total, 21 parallel representations were employed (i.e., $m=21$). These spanned five categories, word and sense, topical, sentiment and affect, psychological and pragmatic, and syntax and style. Regarding thresholds, we used $t_w=0.00001$ and $t_c=0.95$. In order to compare the expressive power for human versus GPT with other between-human demographics, we also performed the same analysis within FCE for age (younger versus older authors) and race (authors from Asian versus non-Asian countries). Finally, as an additional comparison, we included the AskAPatient online forum used in several prior studies, where authors were grouped by gender (male versus female) and age (high versus low) \cite{lalor2022benchmarking}.   

\begin{figure}[ht]
\centering
\includegraphics[width=\textwidth]{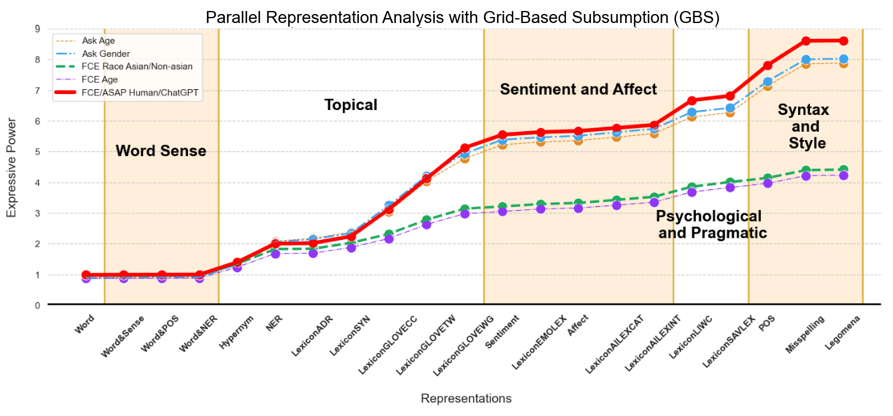}
\caption{Expressive power to demonstrate differences in textual content generated by humans versus GPT}
\label{parallel_img}
\end{figure}

The expressive power results for the five binary class comparisons appear in Figure \ref{parallel_img}, each as a separate line series. Starting with the word representation ($m=1$), assigned an expressive score of 1, going left to right, the chart depicts the cumulative $e(r_j)$ scores associated with each of the additional 20 representations. Looking at the figure, we can see that across word sense representations, all five lines appear similar. However, with the topic-oriented representations (second group from the left), we being to observe differences. Namely, on the GloVe embedding-based lexicons, the human-GPT, Ask-age, and Ask-gender lines show a lift in expressive power, suggesting that the two groups in these three datasets differ in language usage across their topical composition (relative to the FCE-age and FCE-race datasets). This gap between human-GPT and FCE age/race series further widens on the psychological and pragmatic lexicons such as LIWC and speech acts, and again within the syntax and style category (namely for POS tags), suggesting differences in discussion of psychological concepts and actions/intentions appearing in the text. Interestingly, whereas the chart shows increasing linguistic difference between text generated by humans versus GPTs, within the same domain of prompt-driven essays, the differences between human-GPT texts and those observed across age or gender groups in the Ask testbeds is less pronounced. We can conclude that relative to essays written by humans of different groups such as race and age, the GPT generated texts do exhibit greater linguistic variation quantified based on the information gain of tokens across parallel representations. One caveat to this conclusion is that we only have self-reported demographic data for FCE, but not ASAP. On the other hand, the text composition differences when compared against self-reported male/female and young/old authors in an online health discussion forum are much less pronounced. In the ensuing section, we use the attention weights from the BERT models used for assessment to shed further light on these differences between human and GPT text.

\subsection{Content Analysis Using Transformer Attention Weights}
\begin{figure}[ht]
\centering
\includegraphics[width=\textwidth]{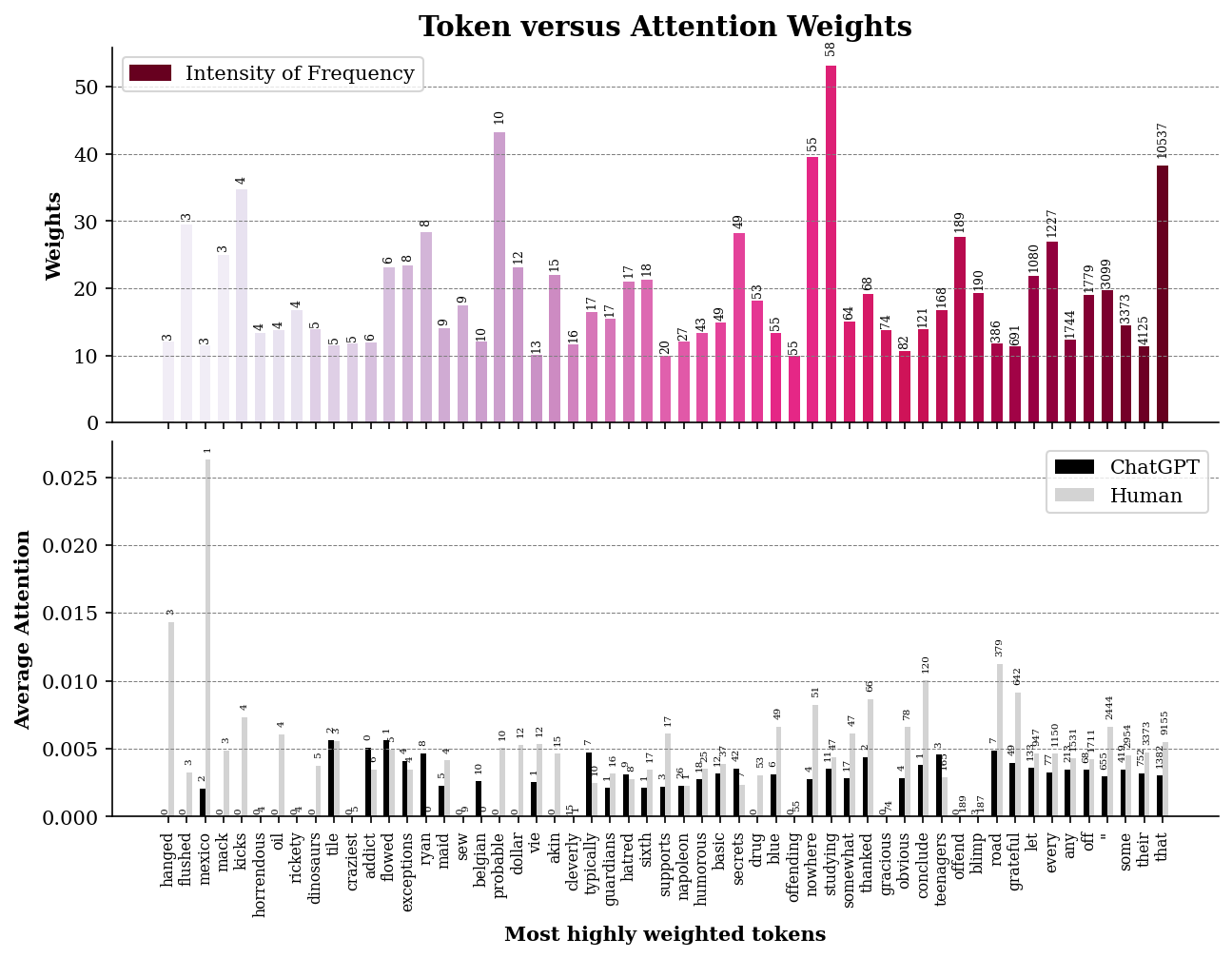}
\caption{Token weights and BERT attention scores for overall highest weighted tokens}
\label{non_exclusive_img}
\end{figure}

Following the approach depicted in Section 3.6, we computed the attention scores $A_i$ for each of the word-representation tokens (i.e., $r_1$) with the highest weights, $w(f_{ij})$. Given our assessment models were trained using 5-fold cross validation, in order to ensure that the attention weights for tokens were only across instances that the model had previously not seen, we computed attention for tokens within a given human-generated text only using the training BERT model from when that text appeared in the test fold. Similarly, for the GPT-generated texts, consistent with the approach taken with our statistical analysis models (Section 3.5), because all texts were out-of-sample, the attention scores for tokens appearing in GPT text were averaged across the 5 training fold models. Figure \ref{non_exclusive_img} shows the results. Each bar represents a token. Token labels appear vertically along the x-axis of the bottom chart. For each token, the top chart depicts token weights along the y-axis along with the occurrence frequency (i.e., how often it appears across the human/GPT essays), with the latter appearing both as the bar label value and color intensity. Tokens are arranged in ascending order from left to right based on their occurrence frequency in the data, which ranges from 3 to 10537. The bottom chart depicts the attention weights (y-axis) and breakdown of occurrences (number labels for each bar) across GPT-generated and human essays (black and gray bars, respectively).

By focusing on the tokens with the highest weights in the word representation, we wanted to see how the BERT model trained on human essays attended to the tokens known to be most skewed in their occurrence for human versus GPT generated text (as measured by their $w(f_{ij})$ weights). Looking at the bottom chart, we can see that with a few exceptions, the attention scores from the BERT model fine-tuned on human essays are almost always higher when those tokens appear in the text. This suggests that because the model's contextualized positional embeddings have been fine-tuned on human essays, it attends more to these tokens when they appear in the text as opposed to when they appear in the out-of-sample GPT essays. Overall, the results from Figure \ref{non_exclusive_img} suggest that the BERT models are attending more to the human text tokens versus the GPT ones. We speculate that this heightened attention could be due to the familiarity in training data, suggesting that the BERT models are attending more to text from contexts they have been fine-tuned on.

\begin{figure}[ht]
\centering
\includegraphics[width=\textwidth]{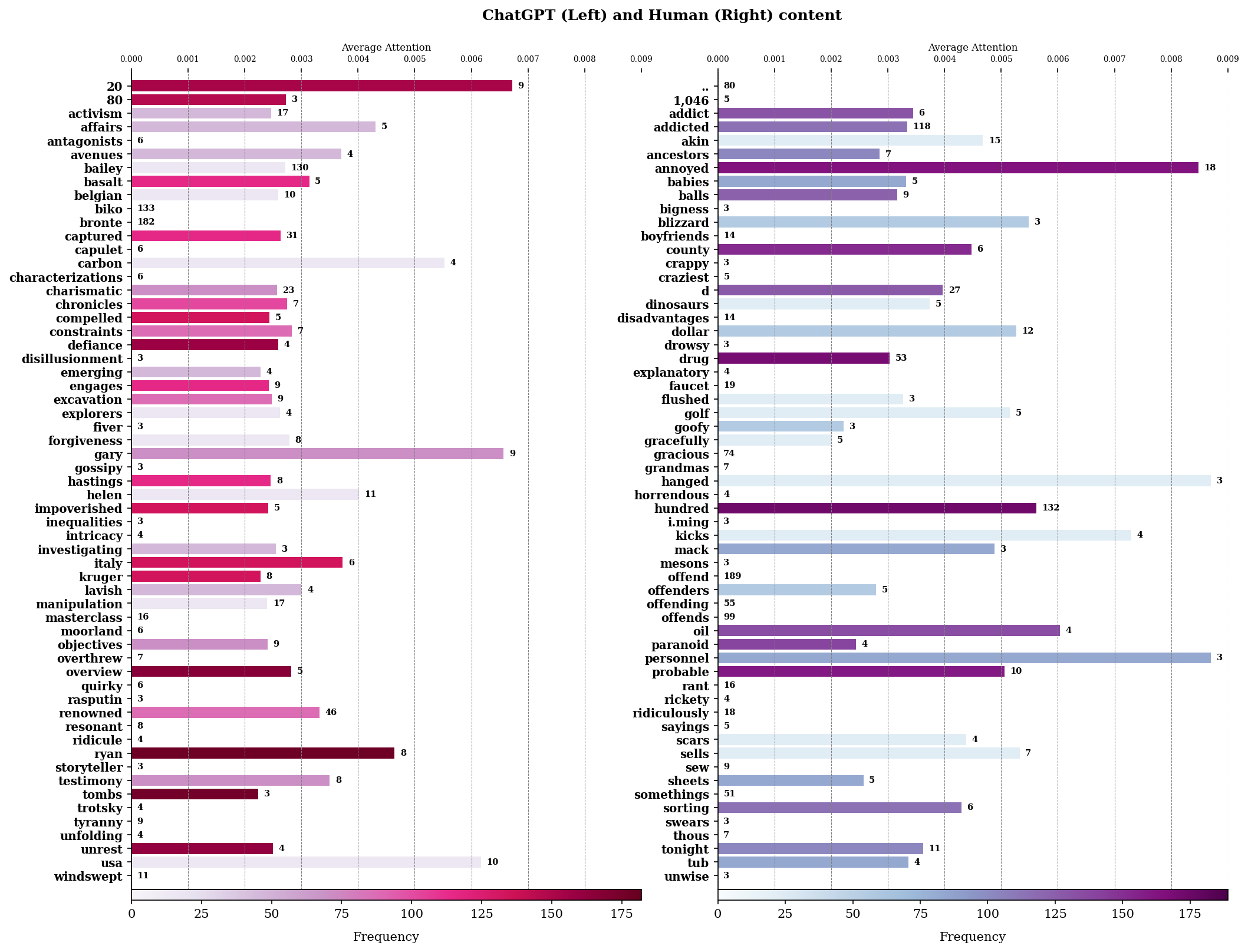}
\caption{Token weights and BERT attention scores for highest weighted tokens only appearing in GPT (left) and human (right) text}
\label{exclusive_img}
\end{figure}

Next, we explored the tokens with the highest weights that only appeared within the GPT or human generated text, respectively. The results appear in Figure \ref{exclusive_img}. The left chart depicts the highest $w(f_{ij})$ tokens from GPT that never appear in human text (and vice versa for the right chart). For each token, the bar shows the average attention scores from BERT (x-axis), and the frequency occurrence of that token in the text (color shade and number appearing next to each bar). Looking at the tokens, we can observe a few interesting patterns. GPT demonstrates far greater use of proper noun names, including authors and characters that are the subject of the essays. Examples include "bronte," "gary," "helen," "bailey" and "hastings." The fact that some/most of these tokens receive attention in the BERT model - when they have not been seen during the fine-tuning process because they never appear in the human text - suggests that the model weights from the corpora that BERT was pre-trained on may be a factor. Admittedly, the positional embeddings and ability of transformer models to pickup on syntactic structure may also be at play - prior studies have noted the complexities of understanding the underlying mechanisms within transformer-based language models \cite{rogers2021primer}. The tokens appearing exclusively in GPT-generated text also include several sophisticated literary concepts and devices, namely, reference to "antagonists," "characterizations," and "intricacy," as well as themes of "forgiveness," "activism," and "manipulation." Conversely, amongst the tokens appearing exclusively in the human-generated text (right chart in Figure \ref{exclusive_img}, there is far greater usage of colloquialisms and less formal verbiage such as "crappy" and "goofy." Moreover, the human text includes far greater usage of tokens with sentiment polarity such as "craziest," "ridiculously," and "horrendous." Collectively, the differences in token occurrences between GPT and human text, as depicted in Figure \ref{exclusive_img} shed light on, and align with, the expressive power trend line presented earlier in Figure \ref{parallel_img} in that the biggest differences between human and GPT text appear across the parts-of-speech (POS), sentiment and affect, and topical concept representations. One notable exception being misspellings, which were naturally more pervasive in human text, but do not manifest in GPT-generated content.

\section{Discussion and Concluding Remarks}
This paper contributes to the nascent yet growing body of literature that explores the intended and unintended consequences of generative AI models. Emerging literature explores the effectiveness of ML generated content directly against a human gold standard \cite{drori2022neural,terwiesch2023would}. In contrast, we explore differences in how ML-based scoring models trained on human content assess the quality of content generated by humans versus GPTs. In particular, we empirically examine how LLMs capable of generating high-quality text may disrupt automated processes that already leverage ML/NLP to score text. Across our three research questions, we firstly show that transformer language model-based assessment methods for scoring text are becoming the state-of-the-art (e.g., BERT and RoBERTa in our study). This is consistent with other text classification problems where models and architectures using robust embeddings are outperforming standard feature ML and CNN/RNN based methods \cite{yang2023getting}. More importantly, related to the second research question, our statistical analysis shows that such transformer language model-based assessment methods rate text generated by LLMs such as GPTs higher than human-generated text, even when they are only trained on human data. For instance, BERT scored GPT-3.5 text about 15\% higher than human text. Notably, we do not observe this trend when the assessment models are feature-based or CNN/RNN. Feature based methods score human text 10-15\% higher whereas this gap is even more pronounced when assessing using CNN/RNN models (26-32\% higher for human text). We explore this disparity for BERT-based models in the content analysis associated with our third research question using parallel representations and attention weights to understand linguistic differences between human and GPT text, and how BERT-based models might be attending differently to GPT versus human text. Our parallel representation analysis shows that there are considerable differences in topics, sentiment, affect, parts-of-speech, and mispellings between GPT and human generated essays. Further, the analysis reveals that the expressive power of the differences is greater than those observed within human texts for cross-gender and cross-race comparisons, but that are on par with differences across gender and age groups for user-generated content from an online discussion forum. Attention weight analysis suggests that BERT models for assessing GPT text are attending to author/character name proper nouns and other tokens that they may have seen during pre-training (e.g., Wikipedia), but that are out-of-sample during fine-tune training on human essays.

Our work presents an important foray into the interplay between text assessment and generative models in the era of LLMs. Our main contributions are two-fold. First, we propose an analysis framework for environments involving automated generation and assessment artifacts, including ML-based scoring models, GPT-generated texts, and statistical models for parsimoniously evaluating such intersections. We intend to make the framework code, analysis models, generated data, and prompt design processes publicly available \footnote{Code and data available at: github.com/nd-hal/automated-ML-scoring-versus-generation}. Second, we use the problem context of AES to offer empirical insights, including how state-of-the-art text scoring methods, based on transformer language models, may score certain genres of GPT-generated content higher even when exclusively trained on human content, possibly due to familiarity. That is, overlap in language model pre-training data between BERT/RoBERTa and GPT-3.5 and GPT-4 could potentially impact how the former assess text generated by the latter. Although set in the context of AES, our framework
and results have implications for many IR and text classification settings where automated scoring
of text/documents is likely to be disrupted by generative AI

This work is not without its limitations. We did not manually label/score the GPT essays – this could shed light on the human-perceived quality of text generated by ChatGPT-type LLMs versus humans. However, in our experience, any manual assessment of GPT generated text could be prone to human-automation biases. On a related note, we did not include the GPT-based texts as part of our assessment model training. However, this was partly motivated by our research questions, which were interested in the impact of injecting generative model-based text into an automated assessment process traditionally devoid of such generated content. Another limitation could be that our interplay occurred in the context of automated essay scoring. However, this is an important problem that has received a fair amount of attention from the NLP community \cite{uto2020neural,uto2021review,shibata2022analytic,Ramesh2022-kg}. Moreover, given the genres of text examined, including persuasion, commentary, response, summary, etc. the results are relevant in related areas such as search, content recommendation, and information retrieval more broadly where similar genres of text manifest in blogs, social media, news articles, online reviews, amongst other document modalities. Nonetheless, we believe our study makes important contributions. We are hopeful that future work can take our conceptual research design and analysis framework and extend it to other domains involving human-LLM assessment/generation intersections, such as fake news detection, search relevance assessment, and recommendation, just to name a few.    

\bibliographystyle{plain}
\bibliography{references}
\appendix

\end{document}